\newif\ifconfver
\confvertrue        %declaring conference version u

\ifconfver
    \documentclass[10pt,journal,compsoc]{IEEEtran}
\else
    \documentclass[12pt,journal,draftcls,onecolumn,compsoc]{IEEEtran}
\fi

\usepackage{graphicx}
\usepackage[ruled]{algorithm2e}
\usepackage[tight]{subfigure}
\usepackage{multirow,amsmath,color,epsfig,amsfonts,amssymb,graphics,psfrag,theorem,calc,url,bm,color,cite}
\usepackage{graphicx}  % Written by David Carlisle and Sebastian Rahtz
\usepackage{stfloats}  % Written by Sigitas Tolusis

    \makeatletter
    \def\multilimits@{\bgroup
  \Let@
  \restore@math@cr
  \default@tag
 \baselineskip\fontdimen10 \scriptfont\tw@
 \advance\baselineskip\fontdimen12 \scriptfont\tw@
 \lineskip\thr@@\fontdimen8 \scriptfont\thr@@
 \lineskiplimit\lineskip
 \vbox\bgroup\ialign\bgroup\hfil$\m@th\scriptstyle{##}$\hfil\crcr}
    \def\Sb{_\multilimits@}
    \def\endSb{\crcr\egroup\egroup\egroup}
\makeatother

{\begin{list}{}%
    {\setlength{\rightmargin}{0pc}%
    \setlength{\leftmargin}{1pc} }} %
{\end{list}}

\newlength{\twidth}
\ifconfver
\else
\fi

\theorembodyfont{\rmfamily}
{\begin{list}{}%
    {\setlength{\rightmargin}{1pc}%
    \setlength{\labelsep}{0pc}
    \setlength{\leftmargin}{1pc} }} %
{\end{list}}

{\begin{list}{}%
    {\setlength{\rightmargin}{0pc}%
    \setlength{\leftmargin}{0pc} }} %
{\end{list}}

{\begin{list}{}{
    \settowidth{\labelwidth}{\mbox{\textnormal{#1}}}%
    \setlength{\leftmargin}{\labelwidth+\labelsep}}}%
{\end{list}}

\newsavebox{\savepar}

\begin{document}
% paper title
\title{PCANet: A Simple Deep Learning Baseline for Image Classification?}

%\ifconfver \else {\linespread{1.1} \rm \fi

\author{Tsung-Han Chan,
        Kui Jia,
        Shenghua Gao,
        Jiwen Lu,
        Zinan Zeng, and
        Yi Ma
\IEEEcompsocitemizethanks{
\IEEEcompsocthanksitem Tsung-Han Chan, Kui Jia, Shenghua Gao, Jiwen Lu, and Zinan Zeng are with Advanced Digital Sciences Center (ADSC), Singapore.
\IEEEcompsocthanksitem Yi Ma is with the School of Information Science and Technology of ShanghaiTech University, and with the ECE Department of the University of Illinois at Urbana-Champaign}
}

\markboth{Journal of \LaTeX\ Class Files}%
{Shell \MakeLowercase{\textit{et al.}}: Bare Demo of IEEEtran.cls for Computer Society Journals}

% make the title area

%\ifconfver \else
%\begin{center} \\[2\baselineskip] \end{center}
%\fi

\IEEEtitleabstractindextext{
\begin{abstract}
%Image classification poses a significant challenge due to a large amount of complex intra-class variability. Learning multi-level (or deep) features for annihilating these variability has recently attracted progressive attention. Nevertheless, training a deep network requires significant domain expertise and effort on adjusting network parameters.
In this work, we propose a very simple deep learning network for image classification which comprises only the very basic data processing components: cascaded principal component analysis (PCA), binary hashing, and block-wise histograms. In the proposed architecture, PCA is employed to learn multistage filter banks. It is followed by simple binary hashing and block histograms for indexing and pooling. This architecture is thus named as a PCA network (PCANet) and can be designed and learned extremely easily and efficiently. For comparison and better understanding, we also introduce and study two simple variations to the PCANet, namely the RandNet and LDANet. They share the same topology of PCANet  but their cascaded filters are either selected randomly or learned from LDA. We have tested these basic networks extensively on many benchmark visual datasets for different tasks, such as LFW for face verification, MultiPIE, Extended Yale B, AR, FERET datasets for face recognition, as well as MNIST for hand-written digits recognition. Surprisingly, for all tasks, such a seemingly naive PCANet model is on par with the state of the art features, either prefixed, highly hand-crafted or carefully learned (by DNNs). Even more surprisingly, it sets new records for many classification tasks in Extended Yale B, AR, FERET datasets, and MNIST variations. Additional experiments on other public datasets also demonstrate the potential of the PCANet serving as a simple but highly competitive baseline for texture classification and object recognition.
\end{abstract}

}

\maketitle
%\ifconfver \else
%   \vspace{1\baselineskip}
%   {\bfseries EDICS}: SAM-APPL (Applications of sensor \& array multichannel
%   processing), SAM-MCHA (Multichannel processing), SAM-SENS (Remote sensing of the
%   environment).
%\fi

% To allow for easy dual compilation without having to reenter the
% abstract/keywords data, the \IEEEtitleabstractindextext text will
% not be used in maketitle, but will appear (i.e., to be "transported")
% here as \IEEEdisplaynontitleabstractindextext when the compsoc
% or transmag modes are not selected <OR> if conference mode is selected
% - because all conference papers position the abstract like regular
% papers do.
\IEEEdisplaynontitleabstractindextext
% \IEEEdisplaynontitleabstractindextext has no effect when using
% compsoc or transmag under a non-conference mode.

% For peer review papers, you can put extra information on the cover
% page as needed:
% \ifCLASSOPTIONpeerreview
% \begin{center} \bfseries EDICS Category: 3-BBND \end{center}
% \fi
%
% For peerreview papers, this IEEEtran command inserts a page break and
% creates the second title. It will be ignored for other modes.
%\IEEEpeerreviewmaketitle

%\IEEEdisplaynontitleabstractindextext
%ken do some trick with line spacing: end here
%\ifconfver \else } \fi
%
%\ifconfver \else
%\newpage
%\fi
%

%%%%%%%%% ABSTRACT

%%%%%%%%% BODY TEXT
\section{Introduction}
Image classification based on visual content is a very challenging task, largely because there is usually large amount of intra-class variability, arising from different lightings, misalignment, non-rigid deformations, occlusion and corruptions. Numerous efforts have been made to counter the intra-class variability by manually designing low-level features for classification tasks at hand. Representative examples are Gabor features and local binary patterns (LBP) for texture and face classification, and SIFT and HOG features for object recognition. While the low-level features can be hand-crafted with great success for some specific data and tasks, designing effective features for new data and tasks usually requires new domain knowledge since most hand-crafted features cannot be simply adopted to new conditions \cite{Hinton2006, Bengio2013}.

Learning features from the data of interest is considered as a plausible way to remedy the limitation of hand-crafted features. An example of such methods is learning through deep neural networks (DNNs), which draws significant attention recently \cite{Hinton2006}. The idea of deep learning is to discover multiple levels of representation, with the hope that higher-level features represent more abstract semantics of the data. Such abstract representations learned from a deep network are expected to provide more invariance to intra-class variability. One key ingredient for success of deep learning in image classification is the use of convolutional architectures \cite{Goodfellow2013, LeCun1998, Jarrett2009, Bruna2013, Lee2009, Krizhevsky2012, Kayukcuoglu2010, Sifre2013}. A convolutional deep neural network (ConvNet) architecture \cite{LeCun1998, Jarrett2009, Kayukcuoglu2010, Goodfellow2013, Krizhevsky2012} consists of multiple trainable stages stacked on top of each other, followed by a supervised classifier. Each stage generally comprises of ``three layers'' -- a convolutional filter bank layer, a nonlinear processing layer, and a feature pooling layer. To learn a filter bank in each stage of ConvNet, a variety of techniques has been proposed, such as restricted Boltzmann machines (RBM) \cite{Lee2009} and regularized auto-encoders or their variations; see \cite{Bengio2013} for a review and references therein. In general, such a network is typically learned by stochastic gradient descent (SGD) method. However, learning a network useful for classification critically depends on expertise of parameter tuning and some {\em ad hoc} tricks.
%non-trivial optimization techniques and even many heuristic and {\em ad hoc} tricks.

While many variations of deep convolutional networks have been proposed for different vision tasks and their success is usually justified empirically, arguably the first instance that has led to clear mathematical justification is the wavelet scattering networks (ScatNet) \cite{Bruna2013,Sifre2013}. The only difference there is that the convolutional filters in ScatNet are prefixed -- they are simply wavelet operators, hence no learning is needed at all. Somewhat surprisingly, such a pre-fixed filter bank, once utilized in a similar multistage architecture of ConvNet or DNNs, has demonstrated superior performance over ConvNet and DNNs in several challenging vision tasks such as handwritten digit and texture recognition \cite{Bruna2013,Sifre2013}. However, as we will see in this paper, such a prefixed architecture does not generalize so well to tasks such as face recognition where the intra-class variability includes significant illumination change and corruption.

\begin{figure}[tbp]
\centering
\resizebox{1\linewidth}{!}{\includegraphics{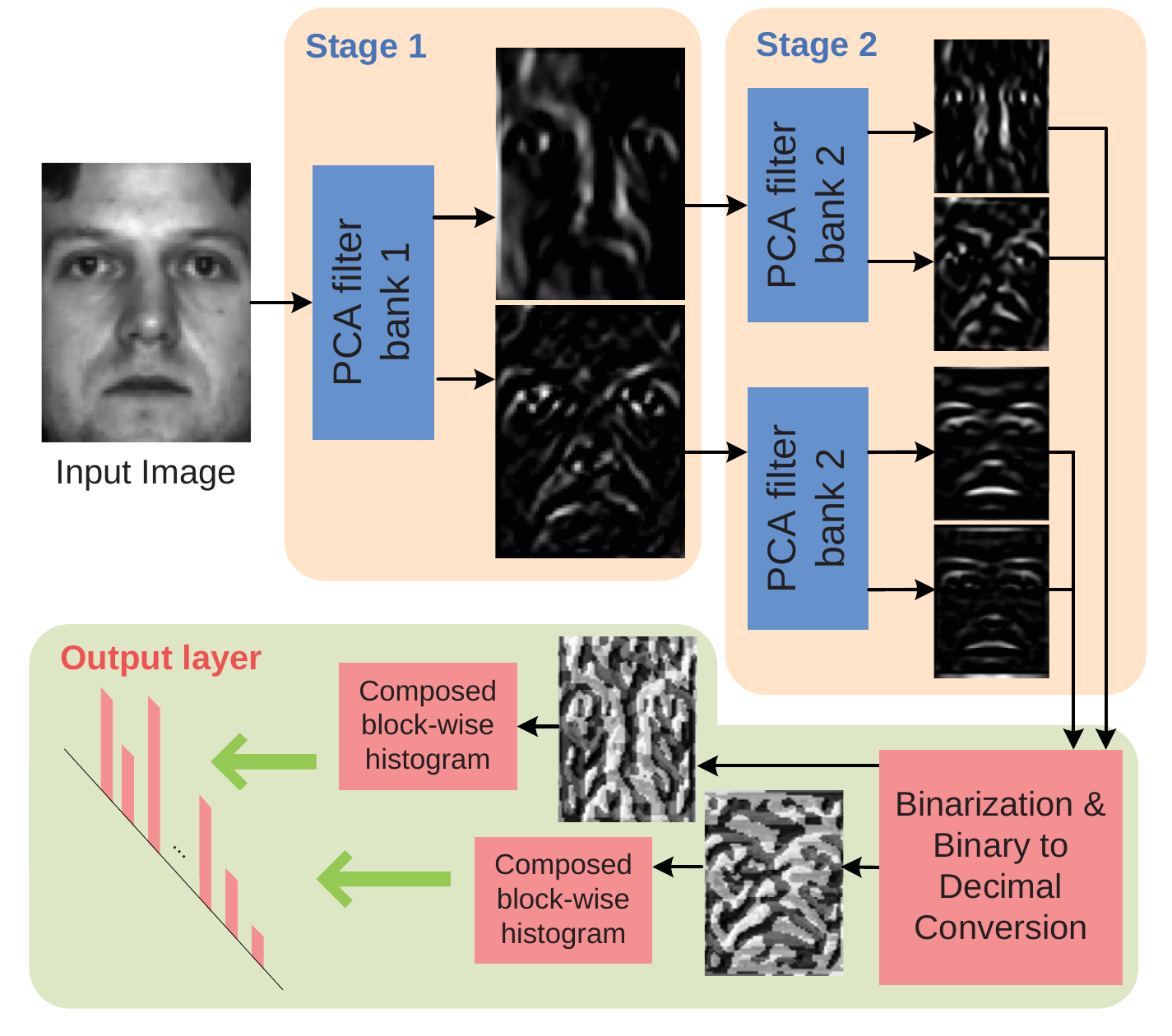}}
\caption{Illustration of how the proposed PCANet extracts features from an image through three simplest processing components: PCA filters, binary hashing, and histogram. } \label{fig: PCANet_concept}%\vspace{-0.5cm}
\vspace{-0.3cm}
\end{figure}

\subsection{Motivations}
An initial motivation of our study is trying to resolve some apparent discrepancies between ConvNet and ScatNet. We want to achieve two simple goals: First, we want to design a simple deep learning network which should be very easy, even trivial, to train and to adapt to different data and tasks. Second, such a basic network could serve as a good baseline for people to empirically justify the use of more advanced processing components or more sophisticated architectures for their deep learning networks.

The solution comes as no surprise: We use the most basic and easy operations to emulate the processing layers in a typical (convolutional) neural network mentioned above: The data-adapting convolution filter bank in each stage is chosen to be the most basic PCA filters; the nonlinear layer is set to be the simplest binary quantization (hashing); for the feature pooling layer, we simply use the block-wise histograms of the binary codes, which is considered as the final output features of the network. For ease of reference, we name such a data-processing network as a {\em PCA Network} (PCANet). As example, Figure \ref{fig: PCANet_concept} illustrates how a two-stage PCANet extracts features from an input image.

At least one characteristic of the PCANet model seem to challenge common wisdoms in building a deep learning network such as ConvNet \cite{LeCun1998, Jarrett2009, Krizhevsky2012} and ScatNet \cite{Bruna2013,Sifre2013}: %1. PCA always gives under-complete representations of the input (image patches) whereas in DNNs, it is usually believed that an over-complete representation in early stages is beneficial;
No nonlinear operations in early stages of the PCANet until the very last output layer where binary hashing and histograms are conducted to compute the output features. Nevertheless, as we will see through extensive experiments, such drastic simplification does not seem to undermine performance of the network on some of the typical datasets.

A network closely related to PCANet could be two-stage oriented PCA (OPCA), which was first proposed for audio processing \cite{Burges2003}. Noticeable differences from PCANet lie in that OPCA does not couple with hashing and local histogram in the output layer. Given covariance of noises, OPCA gains additional robustness to noises and distortions. The baseline PCANet could also incorporate the merit of OPCA, likely offering more invariance to intra-class variability. To this end, we have also explored a supervised extension of PCANet, we replace the PCA filters with filters that are learned from linear discriminant analysis (LDA), called LDANet. As we will see through extensive experiments, the additional discriminative information does not seem to improve performance of the network; see Sections \ref{sec: Extensions} and \ref{sec: experiments}. Another, somewhat extreme, variation to PCANet is to replace the PCA filters with totally random filters (say the filter entries are i.i.d. Gaussian variables), called RandNet. In this work, we conducted extensive experiments and fair comparisons of these types of networks with other existing networks such as ConvNet and ScatNet. We hope our experiments and observations will help people gain better understanding of these different networks.

\subsection{Contributions}
%The proposed PCANet with PCA filters can be trained very efficiently in an unsupervised manner for different types of data (e.g. digits, faces, textures, objects etc.). %The parameters in the PCANet only involve the filter size, the number of stages, the number of filters in each stage, and the block size for the local histogramming.

Although our initial intention of studying the simple PCANet architecture is to have a simple baseline for comparing and justifying other more advanced deep learning components or architectures, our findings lead to some pleasant but thought-provoking surprises: The very basic PCANet, in {\em fair} experimental comparison, is already quite on par with, and often better than, the state-of-the-art features (prefixed, hand-crafted, or learned from DNNs) for almost all image classification tasks, including face images, hand-written digits, texture images, and object images. More specifically, for face recognition with one gallery image per person, it achieves 99.58\% accuracy in Extended Yale B dataset, and over 95\% accuracy for across disguise/illumination subsets in AR dataset. In FERET dataset, it obtains the state-of-the-art average accuracy 97.25\% and achieves the best accuracy of 95.84\% and 94.02\% in Dup-1 and Dup-2 subsets, respectively.\footnote{The results were obtained by following FERET standard training CD, and could be marginally better when the PCANet is trained on MultiPIE database.} In LFW dataset, it achieves competitive 86.28\% face verification accuracy under ``unsupervised setting''. In MNIST datasets, it achieves the state-of-the-art results for subtasks such as basic, background random, and background image. See Section \ref{sec: experiments} for more details. Overwhelming empirical evidences demonstrate the effectiveness of the proposed PCANet in learning robust invariant features for various image classification tasks.

The method hardly contains any deep or new techniques and our study so far is entirely empirical.\footnote{We would be surprised if something similar to PCANet or variations to OPCA \cite{Burges2003} have not been suggested or experimented with before in the vast learning literature.} Nevertheless, a thorough report on such a baseline system has tremendous value to the deep learning and visual recognition community, sending both {\em sobering and encouraging} messages: On one hand, for future study, PCANet can serve as a simple but surprisingly competitive baseline to empirically justify any advanced designs of multistage features or networks. On the other hand, the empirical success of PCANet (even that of RandNet) confirms again certain remarkable benefits from cascaded feature learning or extraction architectures. Even more importantly, since PCANet consists of only a (cascaded) linear map, followed by binary hashing and block histograms, it is now amenable to mathematical analysis and justification of its effectiveness. That could lead to fundamental theoretical insights about general deep networks, which seems in urgent need for deep learning nowadays.

\section{Cascaded Linear Networks}\label{sec: PCANet}

\begin{figure*}[htbp]
\centering
\resizebox{0.9\linewidth}{!}{\includegraphics{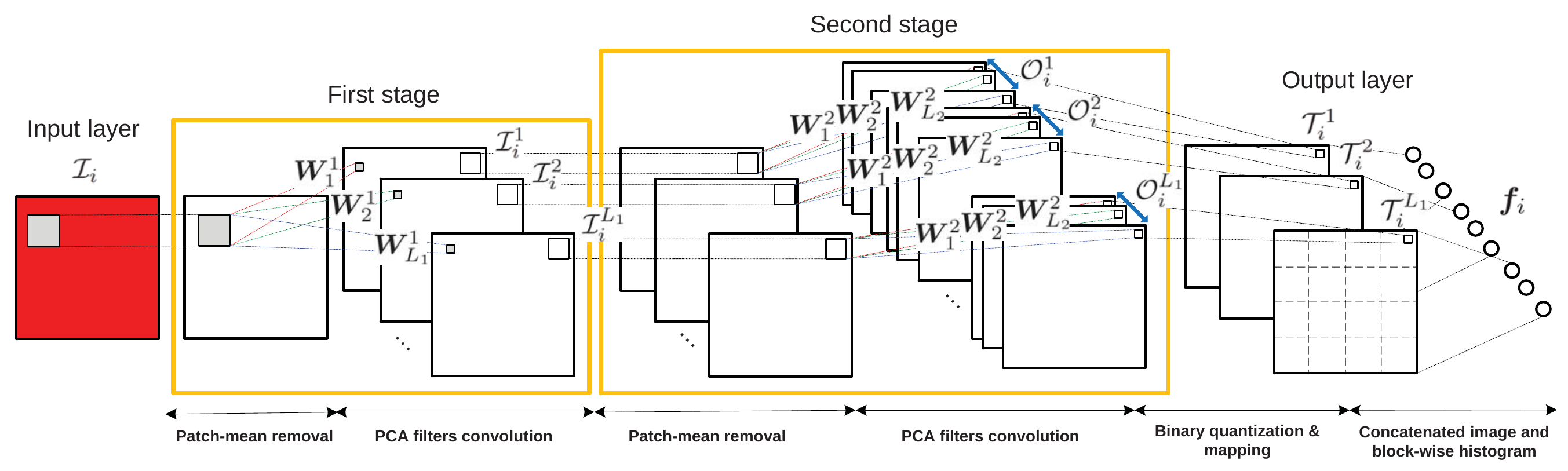}}
\caption{The detailed block diagram of the proposed (two-stage) PCANet.} \label{fig: method1}\vspace{-0.2cm}
\vspace{0cm}
\end{figure*}

\subsection{Structures of the PCA Network (PCANet)}\label{sec: PCANet_main}
Suppose that we are given $N$ input training images $\{{\cal I}_i\}_{i=1}^N$ of size $m\times n$, and assume that the patch size (or 2D filter size) is $k_1 \times k_2$ at all stages. The proposed PCANet model is illustrated in Figure \ref{fig: method1}, and only the PCA filters need to be learned from the input images $\{{\cal I}_i\}_{i=1}^N$. In what follows, we describe each component of the block diagram more precisely.

\subsubsection{\bf The first stage: PCA} \label{sec: The first stage}
Around each pixel, we take a $k_1\times k_2$ patch, and we collect all (overlapping) patches of the $i$th image; i.e., $\bm{ x}_{i,1},\bm{  x}_{i,2},...,\bm{ x}_{i,mn}\in\mathbb{R}^{k_1k_2} $ where each $\bm{ x}_{i,j}$ denotes the $j$th vectorized patch in ${\cal I}_i$. We then subtract patch mean from each patch and obtain $\bar{\bm{ X}}_i = [\bar{\bm{  x}}_{i,1},\bar{\bm{ x}}_{i,2},...,\bar{\bm{ x}}_{i,mn}]$, where $\bar{\bm{ x}}_{i,j}$ is a mean-removed patch. By constructing the same matrix for all input images and putting them together, we get
\begin{equation}\label{eq: datamatrix_1}
\bm{ X} = [\bar{\bm{ X}}_1,\bar{\bm{ X}}_2,...,\bar{\bm{ X}}_N]\in\mathbb{R}^{k_1k_2 \times Nmn}.
\end{equation}Assuming that the number of filters in layer $i$ is $L_i$, PCA minimizes the reconstruction error within a family of orthonormal filters, i.e.,
\begin{equation}
\min_{\bm{ V}\in \mathbb{R}^{k_1k_2 \times L_1}} \|\bm{ X} - \bm{ V}\bm{ V}^T\bm{ X}\|_F^2,~{\rm s.t.}~ \bm{ V}^T\bm{ V} = \bm{  I}_{L_1},
\end{equation}where $\bm{  I}_{L_1}$ is identity matrix of size $L_1\times L_1$. The solution is known as the $L_1$ principal eigenvectors of $\bm{ X}\bm{ X}^T$. The PCA filters are therefore expressed as
\begin{equation}\label{eq: PCAfilter1}
\bm{ W}^1_l \doteq {\rm mat}_{k_1,k_2}(\bm{ q}_l(\bm{ X}\bm{ X}^T))\in\mathbb{R}^{k_1 \times k_2},~l = 1,2,...,L_1,
\end{equation}where ${\rm mat}_{k_1,k_2}(\bm{ v})$ is a function that maps $\bm{ v}\in \mathbb{R}^{k_1k_2}$ to a matrix $\bm{ W}\in \mathbb{R}^{k_1 \times k_2}$, and $\bm{ q}_l(\bm{ X}\bm{ X}^T)$ denotes the $l$th principal eigenvector of $\bm{ X}\bm{ X}^T$. The leading principal eigenvectors capture the main variation of all the mean-removed training patches. Of course, similar to DNN or ScatNet, we can stack multiple stages of PCA filters to extract higher level features.

\subsubsection{\bf The second stage: PCA}
Almost repeating the same process as the first stage. Let the $l$th filter output of the first stage be
\begin{equation}
{\cal I}_i^l \doteq {\cal I}_i*\bm{ W}^1_l,~i = 1,2,...,N,
\end{equation}where $*$ denotes 2D convolution, and the boundary of ${\cal I}_i$ is zero-padded before convolving with $\bm{ W}^1_l$ so as to make ${\cal I}_i^l$ having the same size of ${\cal I}_i$. Like the first stage, we can collect all the overlapping patches of ${\cal I}_i^l$, subtract patch mean from each patch, and form $\bar{\bm{ Y}}^l_i = [\bar{\bm{ y}}_{i,l,1},\bar{\bm{ y}}_{i,l,2},...,\bar{\bm{ y}}_{i,l,mn}]\in\mathbb{R}^{k_1k_2 \times mn}$, where $\bar{\bm{ y}}_{i,l,j}$ is the $j$th mean-removed patch in ${\cal I}_i^l$. We further define $
\bm{ Y}^l = [\bar{\bm{ Y}}^l_1,\bar{\bm{ Y}}^1_2,...,\bar{\bm{ Y}}^l_N]\in\mathbb{R}^{k_1k_2 \times Nmn}$ for the matrix collecting all mean-removed patches of the $l$th filter output, and concatenate $\bm{ Y}^l$ for all the filter outputs as
\begin{equation}
\bm{ Y} = [\bm{ Y}^1,\bm{ Y}^2,...,\bm{ Y}^{L_1}]\in\mathbb{R}^{k_1k_2 \times L_1Nmn}.
\end{equation}
The PCA filters of the second stage are then obtained as
\begin{equation}\label{eq: PCAfilter2}
\bm{ W}^2_{\ell} \doteq {\rm mat}_{k_1,k_2}(\bm{ q}_{\ell}(\bm{ Y}\bm{ Y}^T))\in\mathbb{R}^{k_1 \times k_2},~{\ell} = 1,2,...,L_2.
\end{equation}For each input ${\cal I}_i^l$ of the second stage, we will have $L_2$ outputs, each convolves ${\cal I}_i^l$ with $\bm{ W}^2_{\ell}$ for $\ell = 1,2,...,L_2$:
\begin{equation}
{\cal O}_i^{l} \doteq \{{\cal I}_i^l*\bm{ W}^2_{\ell}\}_{\ell=1}^{L_2}.
\end{equation}The number of outputs of the second stage is $L_1L_2$. One can simply repeat the above process to build more (PCA) stages if a deeper architecture is found to be beneficial.

\subsubsection{\bf Output stage: hashing and histogram}
For each of the $L_1$ input images ${\cal I}_i^l$ for the second stage, it has $L_2$ real-valued outputs $\{{\cal I}_i^l*\bm{ W}^2_{\ell}\}_{\ell=1}^{L_2}$ from the second stage. We binarize these outputs and get $\{H({\cal I}_i^l*\bm{  W}^2_{\ell})\}_{\ell = 1}^{L_2}$, where $H(\cdot)$ is a Heaviside step (like) function whose value is one for positive entries and zero otherwise.

Around each pixel, we view the vector of $L_2$ binary bits as a decimal number. This converts the $L_2$ outputs in ${\cal O}_i^{l}$ back into a single integer-valued ``image'':
\begin{equation}
{\cal T}_i^{l} \doteq \sum_{\ell=1}^{L_2} 2^{\ell-1} H({\cal I}_i^l*\bm{ W}^2_{\ell}),
\end{equation}
whose every pixel is an integer in the range $\big[0,2^{L_2}-1\big]$. The order and weights of for the $L_2$ outputs is irrelevant as we here treat each integer as a distinct ``word.''

For each of the $L_1$ images ${\cal T}_i^l, l = 1, \ldots, L_1$, we partition it into $B$ blocks. We compute the histogram (with $2^{L_2}$ bins) of the decimal values in each block, and concatenate all the $B$ histograms into one vector and denote as ${\rm Bhist}({\cal T}_i^{l})$. After this encoding process, the ``feature'' of the input image ${\cal I}_i$ is then defined to be the set of block-wise histograms; i.e.,
\begin{equation}
\bm{f}_i \doteq [{\rm Bhist}({\cal T}_i^{1}),\ldots,{\rm Bhist}({\cal T}_i^{L_1})]^T \in \mathbb{R}^{(2^{L_2})L_1B}.
\end{equation}
The local blocks can be either overlapping or non-overlapping, depending on applications. Our empirical experience suggests that non-overlapping blocks are suitable for face images, whereas the overlapping blocks are appropriate for hand-written digits, textures, and object images. Furthermore, the histogram offers some degree of translation invariance in the extracted features, as in hand-crafted features (e.g., scale-invariant feature transform (SIFT) \cite{Lowe2004} or histogram of oriented gradients (HOG) \cite{Dalal2005}), learned features (e.g., bag-of-words (BoW) model \cite{Fei2005}), and average or maximum pooling process in ConvNet \cite{LeCun1998, Jarrett2009, Kayukcuoglu2010, Goodfellow2013, Krizhevsky2012}.

The model parameters of PCANet include the filter size $k_1,k_2$, the number of filters in each stage $L_1, L_2$, the number of stages, and the block size for local histograms in the output layer. PCA filter banks require that $k_1k_2 \geq L_1, L_2$. In our experiments in Section \ref{sec: experiments}, we always set $L_1 = L_2 = 8$ inspired from the common setting of Gabor filters \cite{Liu2002} with 8 orientations, although some fine-tuned $L_1, L_2$ could lead to marginal performance improvement. %\footnote{E.g., error rate of PCANet-2 reduces from 1.10 to 1.04 for {\em basic} if $L_1 = 24$.} %Accordingly, $k_1k_2 \geq 8$; an example of minimum square patch is of size $k_1 = k_2 = 3$.
Moreover, we have noticed empirically that two-stage PCANet is in general sufficient to achieve good performance and a deeper architecture does not necessarily lead to further improvement. Also, larger block size for local histograms provides more translation invariance in the extracted feature $\bm{f}_i$.

\subsubsection{Comparison with ConvNet and ScatNet}
Clearly, PCANet shares some similarities with ConvNet \cite{Jarrett2009}. The patch-mean removal in PCANet is reminiscent of local contrast normalization in ConvNet.\footnote{We have tested the PCANet without patch-mean removal and the performance degrades significantly.} This operation moves all the patches to be centered around the origin of the vector space, so that the learned PCA filters can better catch major variations in the data. In addition, PCA can be viewed as the simplest class of auto-encoders, which minimizes reconstruction error.

The PCANet contains no non-linearity process between/in stages, running contrary to the common wisdom of building deep learning networks; e.g., the absolute rectification layer in ConvNet \cite{Jarrett2009} and the modulus layer in ScatNet \cite{Bruna2013,Sifre2013}. We have tested the PCANet with an absolute rectification layer added right after the first stage, but we did not observe any improvement on the final classification results. The reason could be that the use of quantization plus local histogram (in the output layer) already introduces sufficient invariance and robustness in the final feature.

The overall process prior to the output layer in PCANet is completely linear. One may wonder what if we merge the two stages into just one that has an equivalently same number of PCA filters and size of receptive field. To be specific, one may be interested in how the single-stage PCANet with $L_1L_2$ filters of size $(k_1 + k_2 -1) \times (k_1 + k_2 -1)$ could perform, in comparison to the two-stage PCANet we described in Section \ref{sec: PCANet_main}. We have experimented with such settings on faces and handwritten digits and observed that the two-stage PCANet outperforms this single-stage alternative in most cases; see the last rows of Tables \ref{table: MultiPIE}, \ref{table: mnist_standard}, and \ref{table: mnist}. In comparison to the filters learned by the single-stage alternative, the resulting two-stage PCA filters essentially has a low-rank factorization, possibly having lower chance of over-fitting the dataset. As for why we need the deep structure, from a computational perspective, the single-stage alternative requires learning filters with $L_1L_2(k_1 + k_2 -1)^2$ variables, whereas the two-stage PCANet only learns filters with totally $L_1k_1^2 + L_2k_2^2$ variables. Another benefit of the two-stage PCANet is the larger receptive field as it contains more holistic observations of the objects in images and learning invariance from it can essentially capture more semantic information. Our comparative experiments validates that hierarchical architectures with large receptive fields and multiple stacked stages are more efficient in learning semantically related representations, which coincides with what have been observed in \cite{Lee2009}.

{
\subsection{Computational Complexity}
The components for constructing the PCANet are extremely basic and computationally efficient. To see how light the computational complexity of PCANet would be, let us take the two-stage PCANet as an example. In each stage of PCANet, forming the patch-mean-removed matrix $\bm{X}$ costs $k_1k_2 + k_1k_2mn$ flops; the inner product $\bm{X}\bm{X}^T$ has complexity of $2(k_1k_2)^2mn$ flops; and the complexity of eigen-decomposition is ${\cal O}((k_1k_2)^3)$. The PCA filter convolution takes $L_i k_1k_2mn$ flops for stage $i$. In the output layer, the conversion of $L_2$ binary bits to a decimal number costs $2L_2 mn$, and the naive histogram operation is of complexity ${\cal O}(mn B L_2\log 2)$. Assuming $mn \gg \max(k_1,k_2,L_1,L_2, B)$, the overall complexity of PCANet is easy to be verified as $${\cal O}(mn k_1k_2(L_1 + L_2) + mn(k_1k_2)^2).$$
The above computational complexity applies to training phase and testing phase of PCANet, as the extra computational burden in training phase from testing phase is the eigen-decomposition, whose complexity is ignorable when $mn \gg \max(k_1,k_2,L_1,L_2, B)$.

In comparison to ConvNet, the SGD for filter learning is also a simple gradient-based optimization solver, but the overall training time is still much longer than PCANet. For example, training PCANet on around 100,000 images of 80$\times$60 pixel dimension took only half a hour, but CNN-2 took 6 hours, excluding the fine-tuning process; see Section \ref{sec: MultiPIE}.D for details.
}

\subsection{Two Variations: RandNet and LDANet}\label{sec: Extensions}
The PCANet is an extremely simple network, requiring only minimum learning of the filters from the training data. One could immediately think of two possible variations of the PCANet towards two opposite directions:
\begin{enumerate}
\item We could further eliminate the necessity of training data and replace the PCA filters at each layer with random filters of the same size. Be more specific, for random filters, i.e., the elements of $\bm{W}^1_l$ and $\bm{W}^2_l$, are generated following standard Gaussian distribution. We call such a network {\em Random Network}, or RandNet as a shorthand. It is natural to wonder how much degradation such a randomly chosen network would perform in comparison with PCANet.
\item If the task of the learned network is for classification, we could further enhance the supervision of the learned filters by incorporating the information of class labels in the training data and learn the filters based on the idea of multi-class linear discriminant analysis (LDA). We called so composed network {\em LDA Network}, or LDANet for ease of reference. Again we are interested in how much the enhanced supervision would help improve the performance of the network.
\end{enumerate}

To be more clear, we here describe with more details how to construct the LDANet. Suppose that the $N$ training images are classified into $C$ classes $\{{\cal I}_i\}_{i\in S_c}$, $c = 1,2,...,C$ where $S_c$ is the set of indices of images in class $c$, and the mean-removed patches associated with each image of distinct classes $\bar{\bm{X}}_i \in \mathbb{R}^{k_1k_2 \times mn}$, $i \in S_c$ (in the same spirit of $\bar{\bm{X}}_i$ in \eqref{eq: datamatrix_1}) are given. We can first compute the class mean $\bm{\Gamma}_c$ and the intra-class variability $\bm{\Sigma}_c$ for all the patches as follows,
\begin{align}
\bm{\Gamma}_c = & \sum_{i \in S_c} \bar{\bm{X}}_i / |S_c|, \\
\bm{\Sigma}_c = & \sum_{i \in S_c} (\bar{\bm{X}}_i - \bm{\Gamma}_c)(\bar{\bm{X}}_i - \bm{\Gamma}_c)^T / |S_c|.
\end{align}
Each column of $\bm{\Gamma}_c$ denotes the mean of patches around each pixel in the class $c$, and $\bm{\Sigma}_c$ is the sum of all the patch-wise sample covariances in the class $c$. Likewise, the inter-class variability of the patches is defined as
\begin{equation}
\bm{\Phi} = \sum_{c = 1}^C (\bm{\Gamma}_c - \bm{\Gamma})(\bm{\Gamma}_c - \bm{\Gamma})^T / C,
\end{equation}where $\bm{\Gamma}$ is the mean of class means. The idea of LDA is to maximize the ratio of the inter-class variability to sum of the intra-class variability within a family of orthonormal filters; i.e.,
\begin{equation}
\max_{\bm{ V}\in \mathbb{R}^{k_1k_2 \times L_1}} \frac{{\rm Tr}(\bm{V}^T \bm{\Phi} \bm{V})}{{\rm Tr}(\bm{V}^T (\sum_{c = 1}^C\bm{\Sigma}_c) \bm{V})} ,~{\rm s.t.}~ \bm{ V}^T\bm{ V} = \bm{I}_{L_1},
\end{equation}
where ${\rm Tr}(\cdot)$ is the trace operator. The solution is known as the $L_1$ principal eigenvectors of $\tilde{\bm{\Phi}} = (\sum_{c = 1}^C\bm{\Sigma}_c)^{\dag}\bm{\Phi}$, where the superscript ${\dag}$ denotes the pseudo-inverse. The pseudo inverse is to deal with the case when $\sum_{c = 1}^C\bm{\Sigma}_c$ is not of full rank, though there might be another way of handling this with better numeric stability \cite{Yu2001}. The LDA filters are thus expressed as $\bm{W}^1_l = {\rm mat}_{k_1,k_2}(\bm{ q}_l(\tilde{\bm{\Phi}}))\in\mathbb{R}^{k_1 \times k_2},~l = 1,2,...,L_1$. A deeper network can be built by repeating the same process as above. .

\section{Experiments}\label{sec: experiments}
We now evaluate the performance of the proposed PCANet and the two simple variations (RandNet and LDANet) in various tasks, including face recognition, face verification, hand-written digits recognition, texture discrimination, and object recognition in this section.

\subsection{Face Recognition on Many Datasets}
We first focus on the problem of {\it face recognition with one gallery image per person}. We use part of MultiPIE dataset to learn PCA filters in PCANet, and then apply such trained PCANet to extract features of new subjects in MultiPIE dataset, Extended Yale B, AR, and FERET datasets for face recognition.

\subsubsection{Training and Testing on MultiPIE Dataset.}\label{sec: MultiPIE}
{\bf Generic faces training set.} MultiPIE dataset \cite{Gross2008} contains 337 subjects across simultaneous variation in pose, expression, and illumination. Of these 337 subjects, we select the images of 129 subjects that enrolled all the four sessions. The images of a subject under all illuminations and all expressions at pose $-30^\circ$ to $+30^\circ$ with step size $15^\circ$, a total of 5 poses, were collected. We manually select eye corners as the ground truth for registration, and down-sample the images to 80$\times$60 pixels. The distance between the two outer eye corners is normalized to be 50 pixels. This {\em generic faces training set} comprises around 100,000 images, and all images are converted to gray scale.

We use these assembled face images to train the PCANet and together with data labels to learn LDANet, and then apply the trained networks to extract features of the new subjects in Multi-PIE dataset. As mentioned above, 129 subjects enrolling all four sessions are used for PCANet training. The remaining 120 new subjects in Session 1 are used for gallery training and testing. Frontal view of each subject with neutral expression and frontal illumination is used in gallery, and the rest is for testing. We classify all the possible variations into 7 test sets, namely cross illumination, cross expression, cross pose, cross expression-plus-pose, cross illumination-plus-expression, cross illumination-plus-pose, and cross illumination-plus-expression-and-pose. The cross-pose test set is specifically collected over the poses $-30^\circ$, $-15^\circ$, $+15^\circ$, $+30^\circ$.

%Moreover, other than the PCA filters in PCANet, we also employ random filters and supervised filters in the proposed simple network structure. For random filters, their elements are generated following standard Gaussian distribution. The supervised filters are learned by linear discriminant analysis (LDA). We named the former RandNet, and the latter LDANet.

\begin{figure}[t]
\centering
\subfigure[tight][]{
\resizebox{0.45\linewidth}{!}{\includegraphics{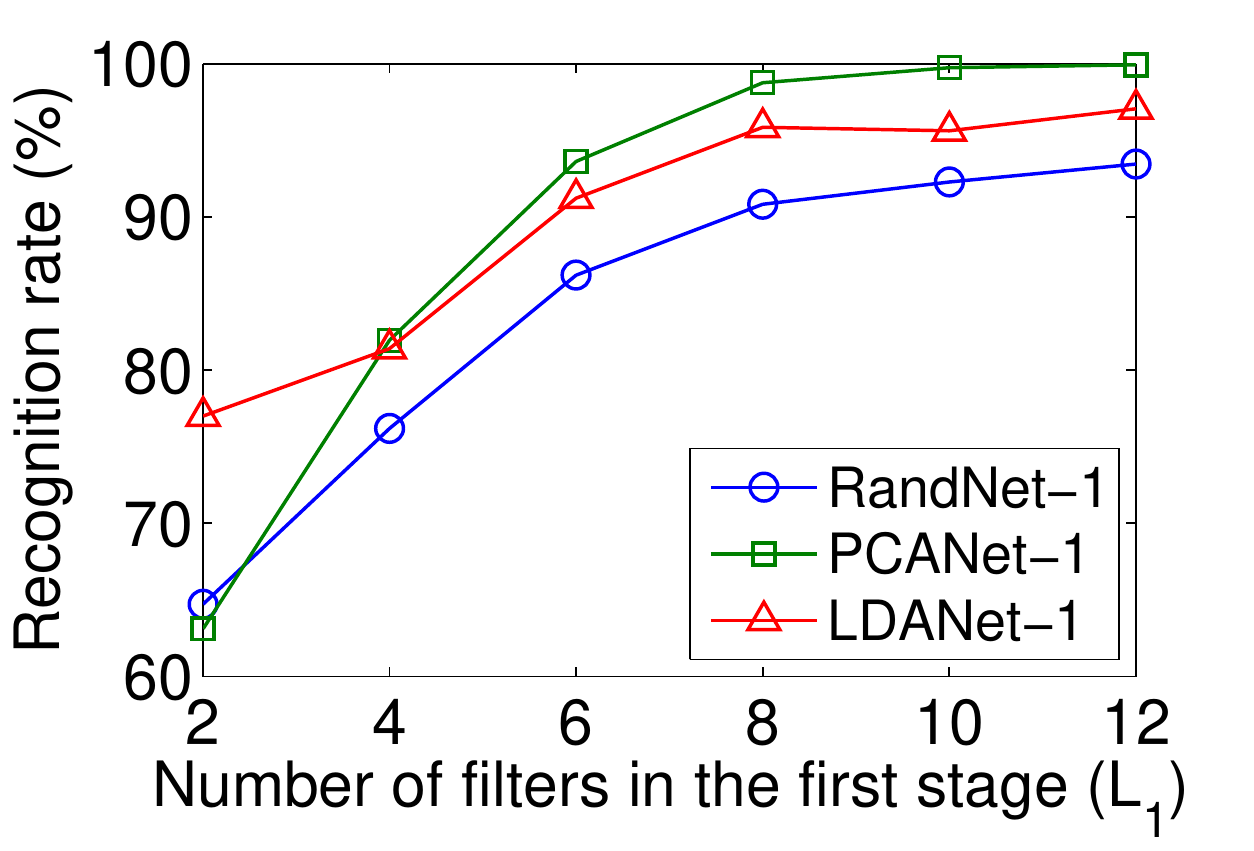}}\label{fig:
MultiPIE_OneStage_K1}}\hspace{0cm} \subfigure[tight][]{
\resizebox{0.45\linewidth}{!}{\includegraphics{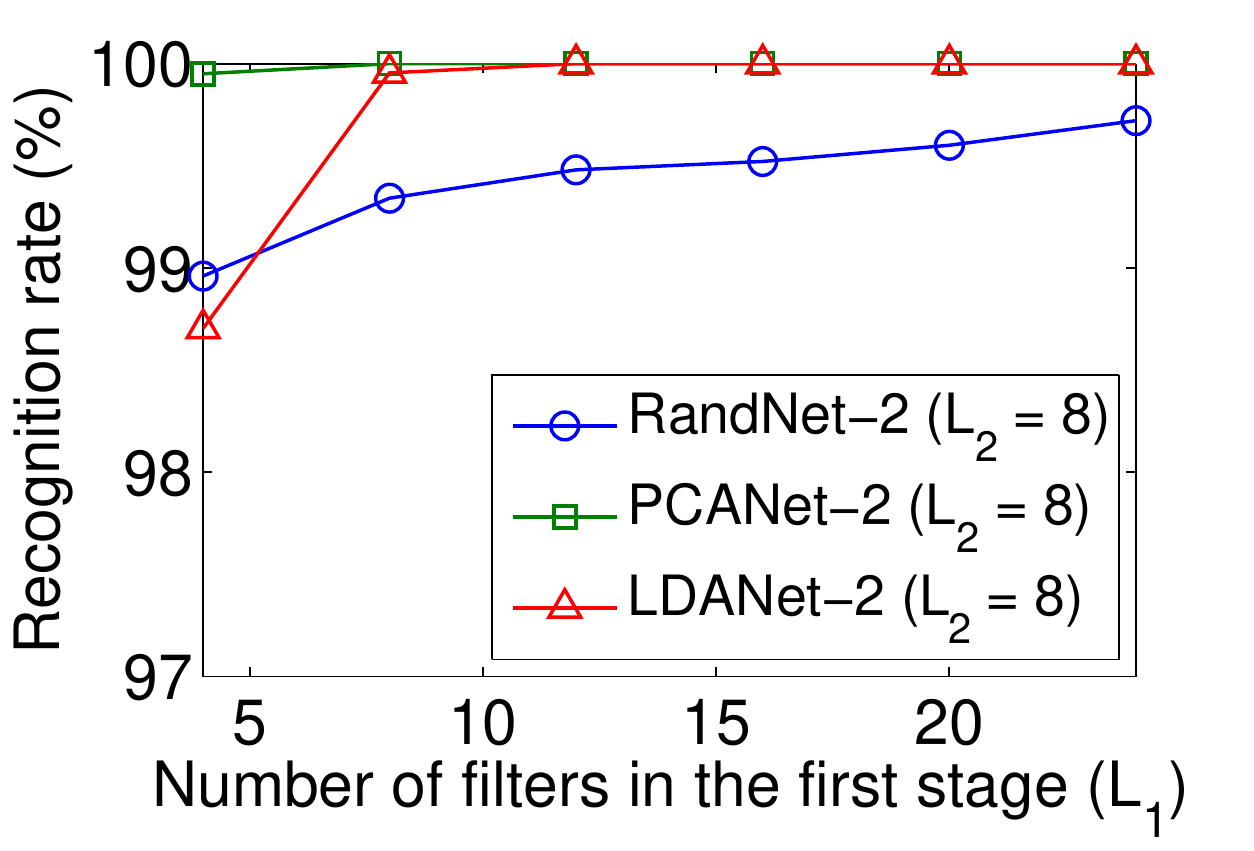}}\label{fig:
MultiPIE_TwoStage_K1}}
\caption{Recognition accuracy of PCANet on MultiPIE cross-illumination test set for varying number of filters in the first stage. (a) PCANet-1; (b) PCANet-2 with $L_2 = 8$.}\label{fig: MultiPIE_K1}
\end{figure}

\vspace{0.3\baselineskip}
{\em A. Impact of the number of filters.} Before comparing RandNet, PCANet, and LDANet with existing methods on all the 7 test sets, we first investigate the impact of the number of filters of these networks on the cross-illumination test set only. The filter size of the networks is $k_1 = k_2 = 5$ and their non-overlapping blocks is of size 8$\times$6. We vary the number of filters in the first stage $L_1$ from $2$ to $12$ for one-stage networks. When considering two-stage networks, we set $L_2=8$ and vary $L_1$ from $4$ to $24$. The results are shown in Figure \ref{fig: MultiPIE_K1}. One can see that PCANet-1 achieves the best results for $L_1 \geq 4$ and PCANet-2 is the best for all $L_1$ under test. Moreover, the accuracy of PCANet and LDANet (for both one-stage and two-stage networks) increases for larger $L_1$, and the RandNet also has similar performance trend. However, some performance fluctuation is observed for RandNet due to the filters' randomness.

\begin{figure}[t]
\centering
\resizebox{0.85\linewidth}{!}{\includegraphics{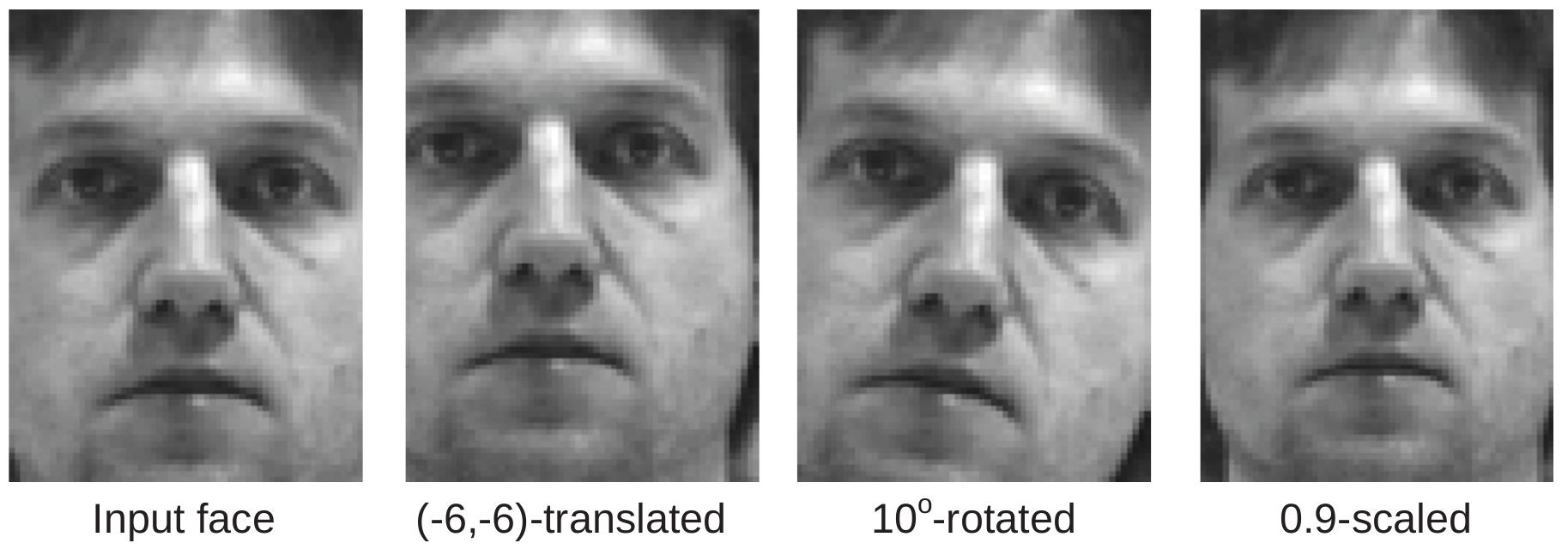}}
\caption{Original image and its artificially deformed images.}\label{fig: artificial}
\end{figure}

\begin{figure*}[t]
\centering
%\begin{minipage}[b]{\linewidth}
\subfigure[tight][]{
\resizebox{0.7\linewidth}{!}{\includegraphics{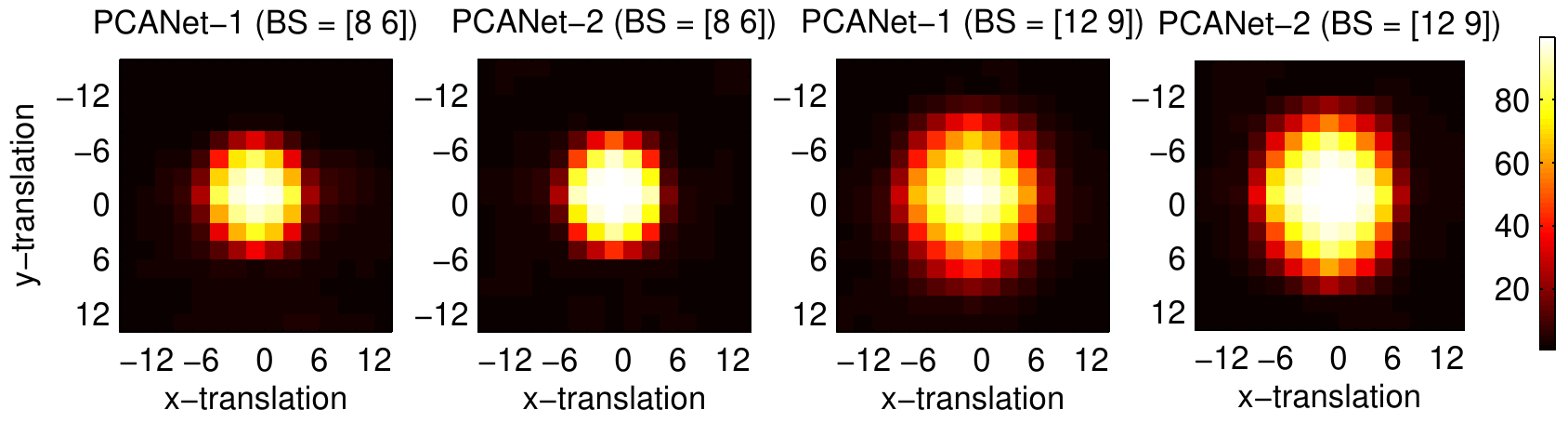}}\label{fig:
MultiPIE_translation}}
%\\
\quad \quad \quad \quad
\subfigure[tight][]{
\resizebox{0.23\linewidth}{!}{\includegraphics{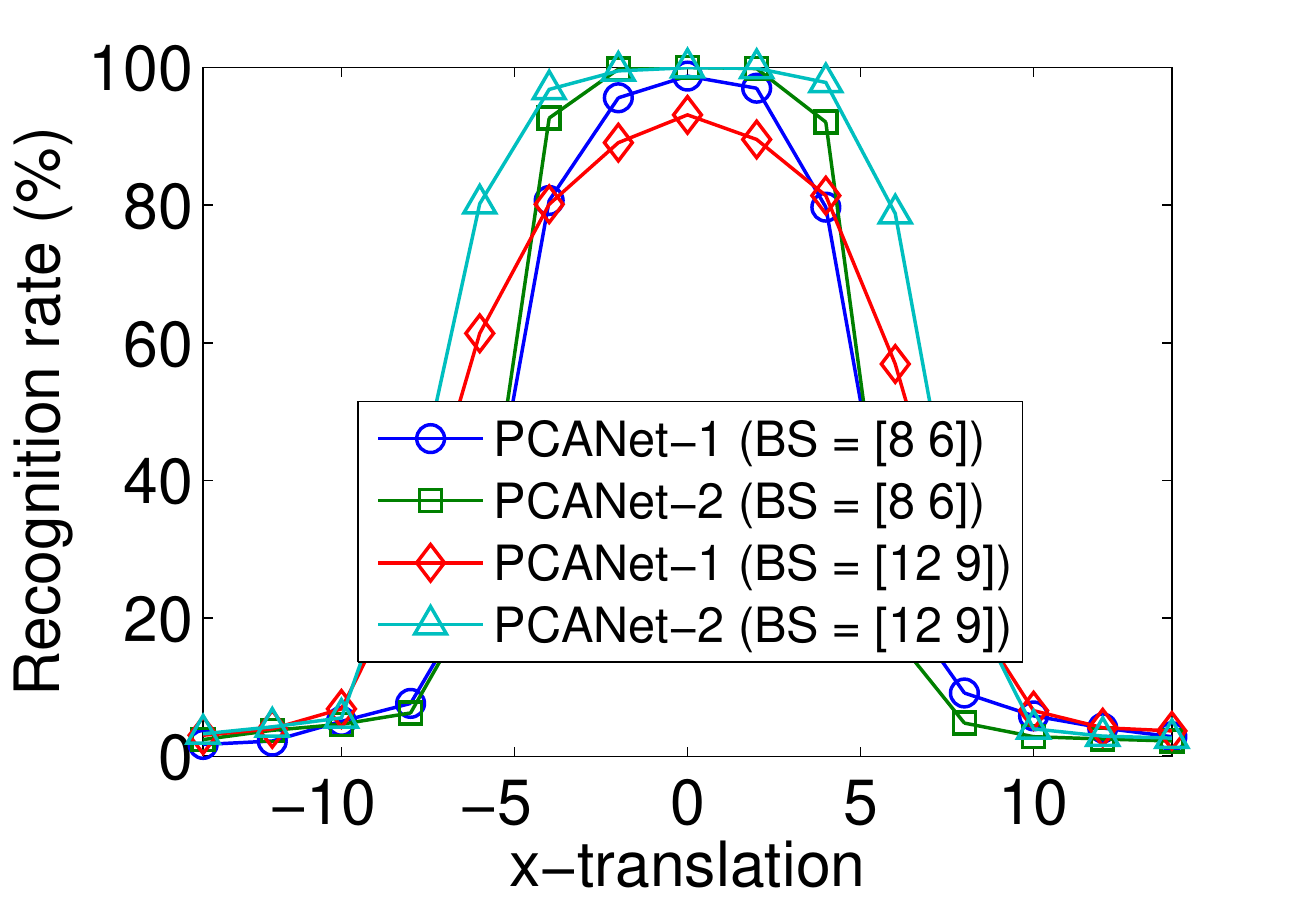}}\label{fig:
MultiPIE_translation_x}}\hspace{0cm}
\subfigure[tight][]{
\resizebox{0.23\linewidth}{!}{\includegraphics{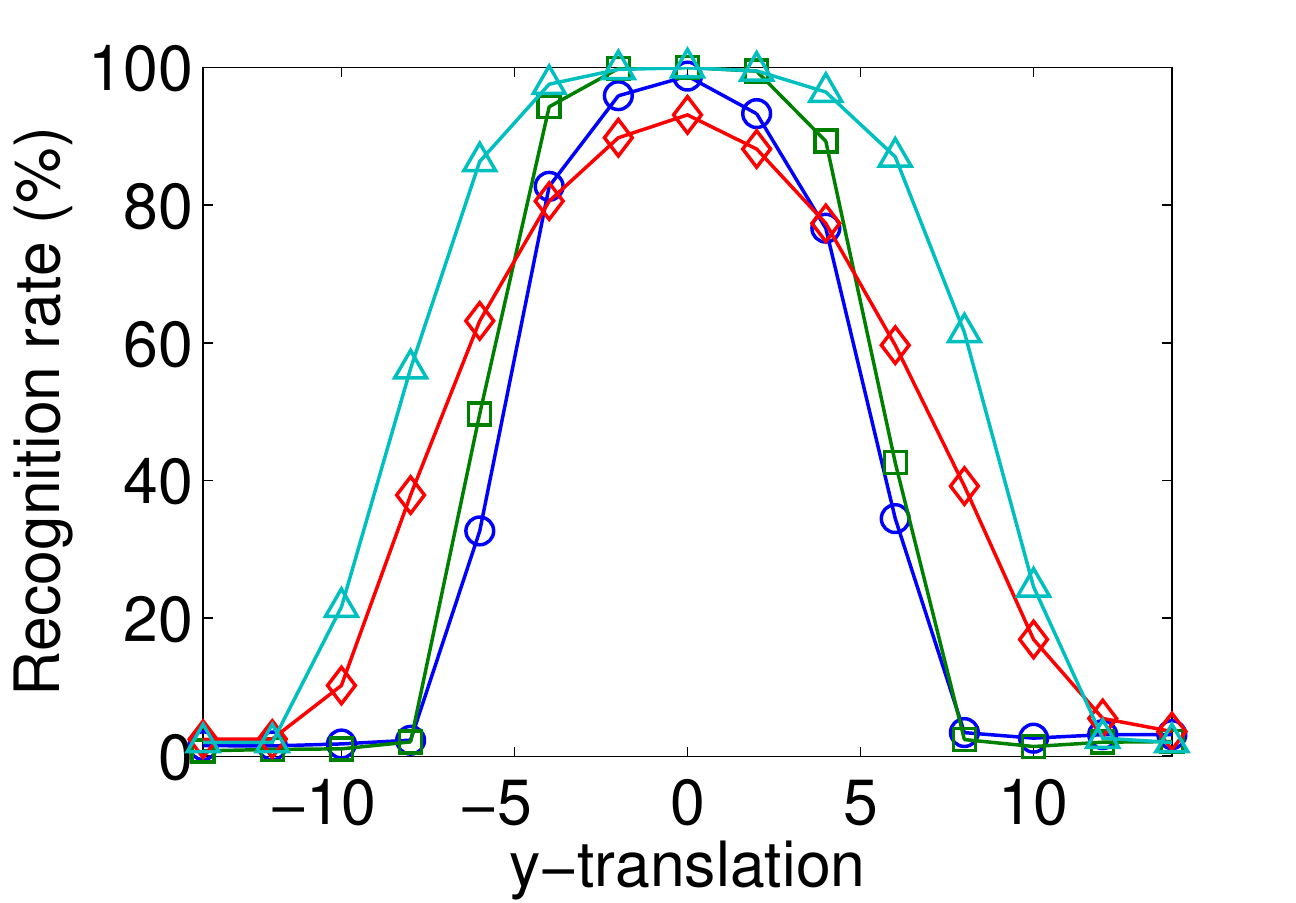}}\label{fig:
MultiPIE_translaion_y}}
\subfigure[tight][]{
\resizebox{0.23\linewidth}{!}{\includegraphics{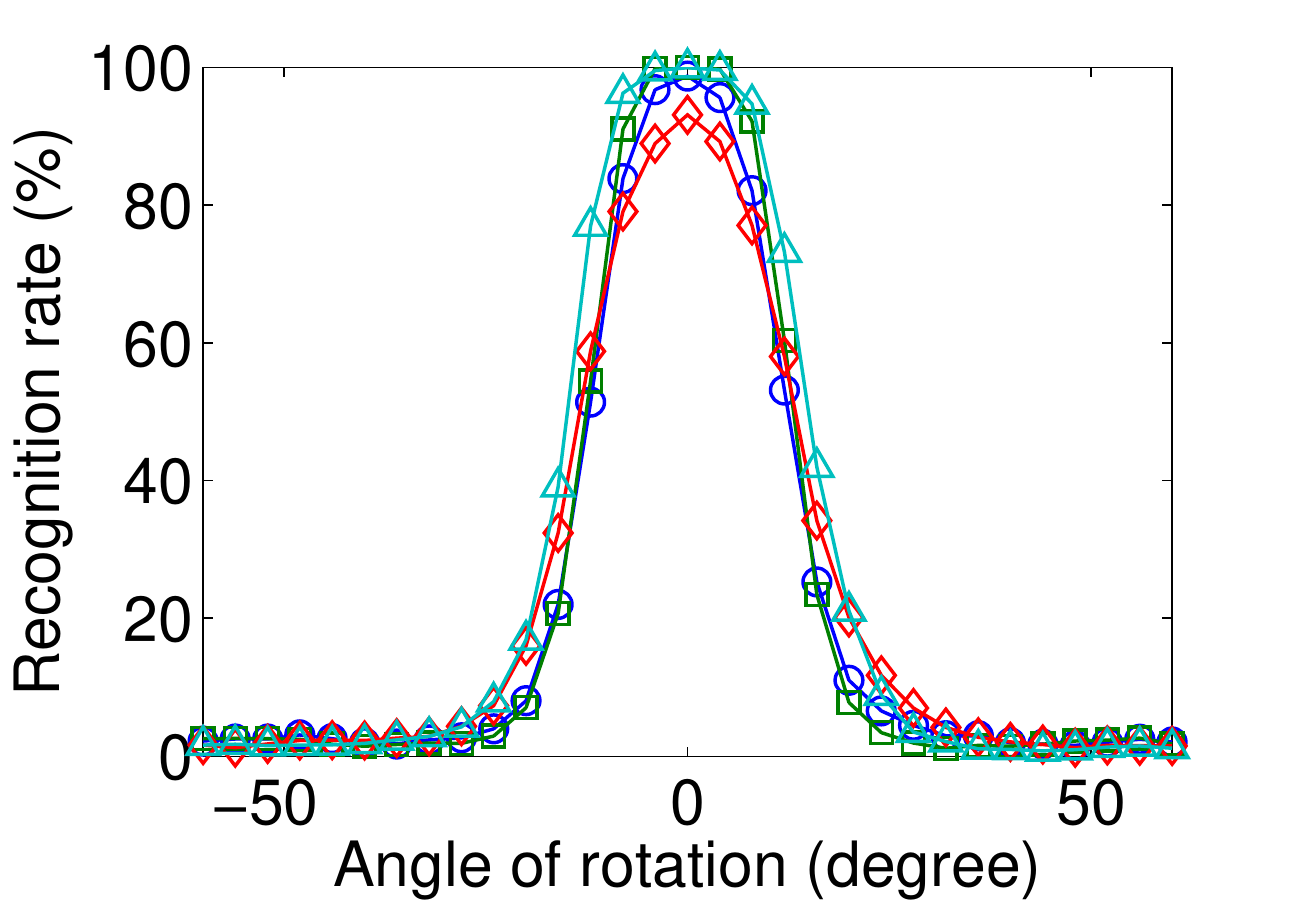}}\label{fig:
MultiPIE_rotation}}\hspace{0cm}
\subfigure[tight][]{
\resizebox{0.23\linewidth}{!}{\includegraphics{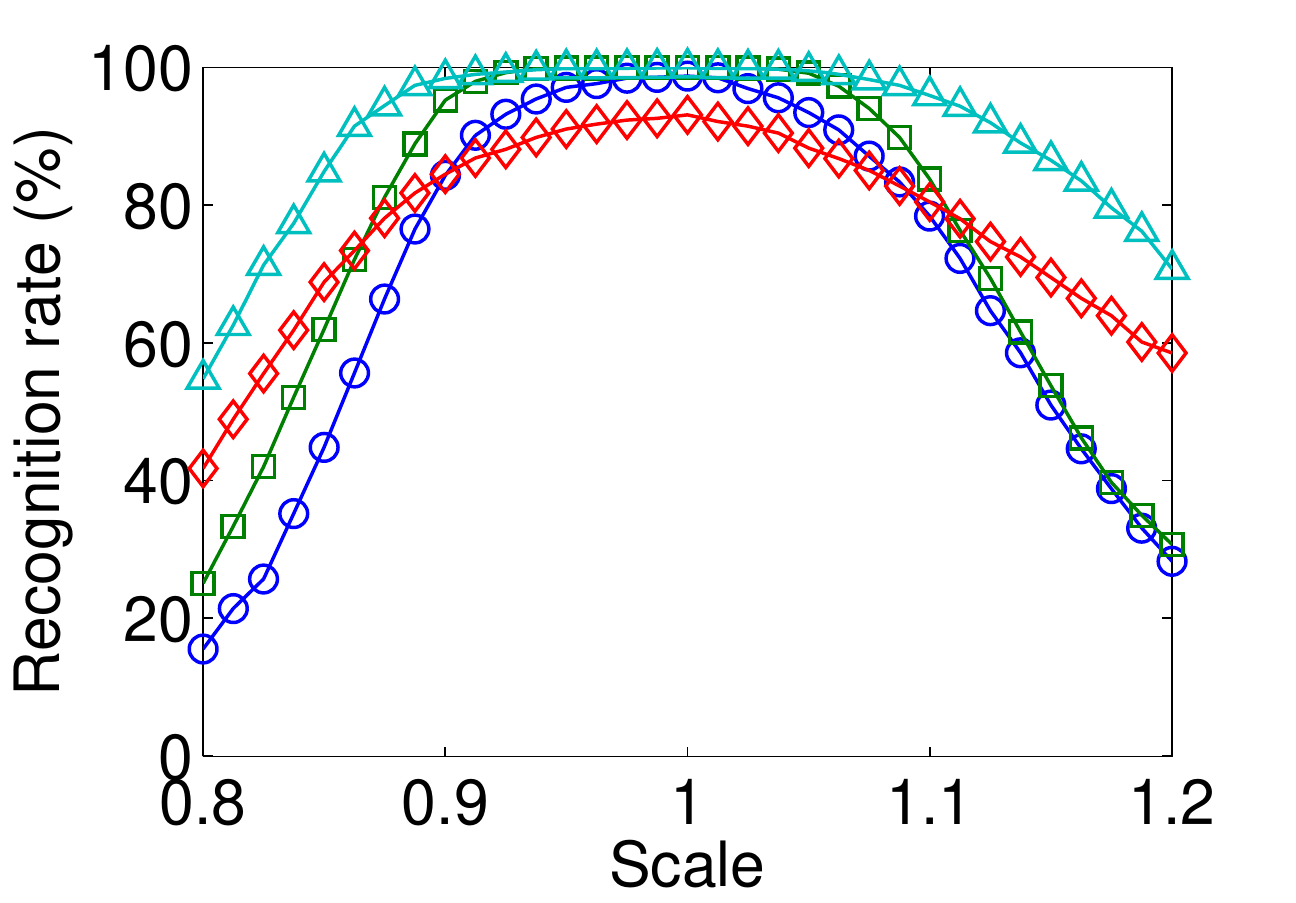}}\label{fig:
MultiPIE_scale}}

%\end{minipage}
\caption{Recognition rate of PCANet on MultiPIE cross-illumination test set, for different PCANet block size and deformation to the test image. Two block sizes [8 6] and [12 9] for histogram aggregation are tested. (a) Simultaneous translation in $x$ and $y$ directions. (b) Translation in $x$ direction. (c) Translation in $y$ direction. (d) In-plane rotation. (e) Scale variation. }\label{fig: MultiPIE_misalignment}
\end{figure*}

\vspace{0.3\baselineskip}
{\em B. Impact of the the block size.} We next examine the impact of the block size (for histogram computation) on robustness of PCANet against image deformations. We use the cross-illumination test set, and introduce artificial deformation to the testing image with a translation, in-plane rotation or scaling; see Figure \ref{fig: artificial}. The parameters of PCANet are set to $k_1 = k_2 = 5$ and $L_1 = L_2 = 8$. Two block sizes 8$\times$6 and 12$\times$9 are considered. Figure \ref{fig: MultiPIE_misalignment} shows the recognition accuracy for each artificial deformation. It is observed that PCANet-2 achieves more than 90 percent accuracy with translation up to 4 pixels in all directions, up to 8$^\circ$ in-plane rotation, or with scale varying from 0.9 to 1.075. Moreover, the results suggest that PCANet-2 with larger block size provides more robustness against various deformations, but a larger block side may sacrifice some performance for PCANet-1.

\vspace{0.3\baselineskip}
{\em C. Impact of the number of generic faces training samples.} We also report the recognition accuracy of the PCANet for differen number of the generic faces training images. Again, we use cross-illumination test set. We randomly select $S$ images from the generic training set to train the PCANet, and varies $S$ from $10$ to $50,000$. The parameters of PCANet are set to $k_1 = k_2 = 5$, $L_1 = L_2 = 8$, and block size 8$\times$6. The results are tabulated in Table \ref{table: MultiPIE_generic_training}. Surprisingly, the accuracy of PCANet is less-sensitive to the number of generic training images. The performance of PCANet-1 gradually improves as the number of generic training samples increases, and PCANet-2 keeps perfect recognition even when there are only 100 generic training samples.

\begin{table}\centering
\caption{Face recognition rates $(\%)$ of PCANet on MultiPIE cross-illumination test set, with respect to different amount of generic faces training images ($S$). }
\vspace{0.1cm}\begin{tabular}{c|c|c|c|c|c|c}
  \hline
  % after \\: \hline or \cline{col1-col2} \cline{col3-col4} ...
  $S$ & 100 & 500 & 1,000 & 5,000 & 10,000 & 50,000  \\ \hline \hline
  PCANet-1   & 98.01 & 98.44 & 98.61 & 98.65 & 98.70 & 98.70  \\
  PCANet-2   & 100.00 & 100.00 & 100.00 & 100.00 & 100.00 & 100.00 \\
  \hline
\end{tabular}\label{table: MultiPIE_generic_training}
\end{table}

\begin{figure}[t]
\centering
\resizebox{0.9\linewidth}{!}{\includegraphics{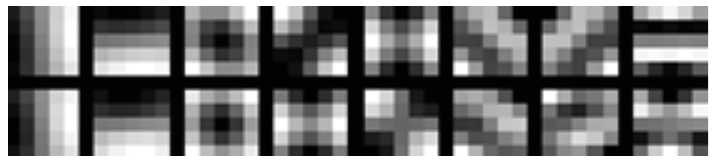}}
\caption{The PCANet filters learned on MultiPIE dataset. Top row: the first stage. Bottom row: the second stage.} \label{fig: MultiPIE_filters}
\end{figure}

\vspace{0.3\baselineskip}
{\em D. Comparison with state of the arts.} We compare the RandNet, PCANet, and LDANet with Gabor\footnote{Each face is convolved with a family of Gabor kernels with 5 scales and 8 orientations. Each filter response is down-sampled by a $3\times 3$ uniform lattice, and normalized to zero mean and unit variance.} \cite{Liu2002}, LBP\footnote{Each face is divided into several blocks, each of size the same as PCANet. The histogram of 59 uniform binary patterns is then computed, where the patterns are generated by thresholding 8 neighboring pixels in a circle of radius 2 using the central pixel value.} \cite{Ahonen2006}, and two-stage ScatNet (ScatNet-2) \cite{Bruna2013}. We set the parameters of PCANet to the filter size $k_1 = k_2 = 5$, the number of filters $L_1 = L_2 = 8$, and 8$\times$6 block size, and the learned PCANet filters are shown in Figure \ref{fig: MultiPIE_filters}. The number of scales and the number of orientations in ScatNet-2 are set to 3 and 8, respectively. We use the nearest neighbor (NN) classifier with the chi-square distance for RandNet, PCANet, LDANet and LBP, or with the cosine distance for Gabor and ScatNet. The NN classifier with different distance measure is to secure the best performances of respective features.

We also compare with CNN. Since we could not find any work that successfully applies CNN to the same face recognition tasks, we use Caffe framework \cite{Jia13caffe} to pre-train a two-stage CNN (CNN-2) on the generic faces training set. The CNN-2 is fully-supervised network with filter size 5$\times$5; 20 channels for the first stage and 50 channels for the second stage. Each convolution output is followed by a rectified linear function $relu(x) = \max(x,0)$ and 2$\times$2 max-pooling. The output layer is a softmax classifier. After pre-training the CNN-2 on the generic faces training set, the CNN-2 is also fine-tuned on the 120 gallery images for $500$ epochs.

The performance of all methods are given in Table \ref{table: MultiPIE}. Except for cross-pose test set, the PCANet yields the best precision. For all test sets, the performance of RandNet and LDANet is inferior to that of PCANet, and LDANet does not seem to take advantage of discriminative information. One can also see that whenever there is illumination variation, the performance of LBP drops significantly. The PCANet overcomes this drawback and offers comparable performance to LBP for cross-pose and cross-expression variations. As a final note, ScatNet and CNN seem not performing well.\footnote{The performance of CNN could be further promoted if the model parameters are more fine-tuned.} This is the case for all face-related experiments below, and therefore ScatNet and CNN are not included for comparison in these experiments. We also do not include RandNet and LDANet in the following face-related experiments, as they did not show performance superior over PCANet.

The last row of Table \ref{table: MultiPIE} shows the result of PCANet-1 with $L_1L_2$ filters of size $(k_1 + k_2 -1) \times (k_1 + k_2 -1)$. The PCANet-1 with such a parameter setting is to mimic the reported PCANet-2 in a single-stage network, as both have the same number of PCA filters and size of receptive field. PCANet-2 outperforms the PCANet-1 alternative, showing the advantages of deeper networks. Another issue worth mentioning is the efficiency of the PCANet. Training PCANet-2 on the generic faces training set (i.e., around 100,000 face images of 80$\times$60 pixel dimension) took only half a hour, but CNN-2 took 6 hours, excluding the fine-tuning process.

\begin{table*}\centering
\caption{Comparison of face recognition rates $(\%)$ of various methods on MultiPIE test sets. The filter size $k_1 = k_2 = 5$ are set in RandNet, PCANet, and LDANet unless specified otherwise.}
\vspace{0.1cm}\begin{tabular}{l|c|c|c|c|c|c|c}
  \hline
  % after \\: \hline or \cline{col1-col2} \cline{col3-col4} ...
  Test Sets & Illum. & Exps. & Pose & Exps.+Pose & Illum.+Exps. & Illum.+Pose & Illum.+Exps.+Pose \\ \hline \hline
  Gabor \cite{Liu2002}   & 68.75  & 94.17 & 84.17 & 64.70 & 38.09 & 39.76 & 25.92 \\
  LBP  \cite{Ahonen2006}   & 79.77 & 98.33 & {\bf 95.63} & 86.88 & 53.77 & 50.72 & 40.55 \\
  ScatNet-2 \cite{Bruna2013} & 20.88 & 66.67 & 71.46 & 54.37 & 14.51 & 15.00 & 14.47 \\
  CNN-2 \cite{Krizhevsky2012} & 46.71 & 75.00 & 73.54 & 57.50 & 23.38 & 25.05 & 18.74 \\  \hline
%  Pc-LBP  & 98.84 & {\bf 100} & 86.46 & 74.17 & 70.10 & 57.73 & 45.87 \\
  RandNet-1 &    80.88  &    98.33   &    87.50     &     75.62   &       46.57     &     42.80    &       31.85 \\
  RandNet-2 &      97.64  &    97.50   &    83.13    &      75.21    &      63.87    &      53.50  &         42.47 \\
  PCANet-1   & 98.70 & {\bf 99.17} & 94.17 & {\bf 87.71} & 72.40 & 65.76 & 53.80 \\
  PCANet-2   & {\bf 100.00} & {\bf  99.17} & 93.33 & 87.29 & {\bf 87.89} & {\bf 75.29} & {\bf 66.49} \\
  LDANet-1   &  99.95  &    98.33   &    92.08   &       82.71     &     77.89   &       68.55    &       57.97    \\
  LDANet-2   & 96.02  &    99.17  &     93.33    &      83.96    &      65.78    &      60.14      &     46.72    \\ \hline
  PCANet-1 ($k_1 = 9$) & {\bf 100}  & {\bf 99.17} & 89.58 &  81.46 & 75.74 & 67.59 & 56.95  \\
  \hline
\end{tabular}\label{table: MultiPIE}
\end{table*}

\subsubsection{Testing on Extended Yale B Dataset.}\label{sec: ExtendedYaleB}
\begin{figure}[t]
\centering
\resizebox{0.85\linewidth}{!}{\includegraphics{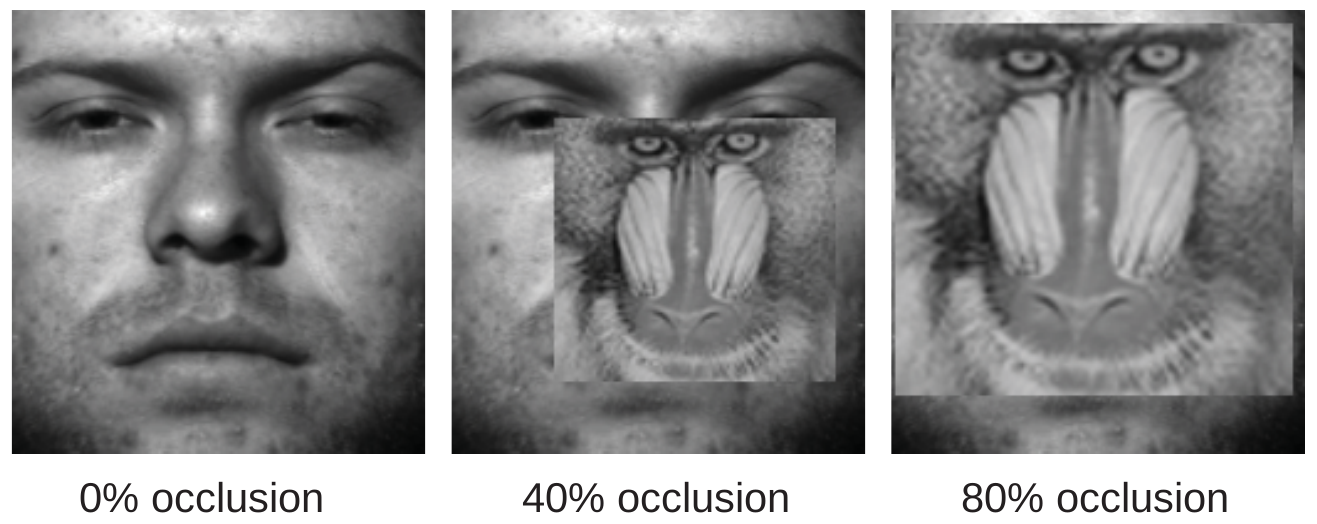}}
\caption{Illustration of varying level of an occluded test face image.}\label{fig: EYaleB_occlusion}
\end{figure}

We then apply the PCANet with the PCA filters learned from MultiPIE to Extended Yale B dataset \cite{Georghiades2001}. Extended Yale B dataset consists of 2414 frontal-face images of 38 individuals. The cropped and normalized 192$\times$168 face images were captured under various laboratory-controlled lighting conditions. For each subject, we select frontal illumination as the gallery images, and the rest for testing. To challenge ourselves, in the test images, we also simulate various levels of contiguous occlusion, from 0 percent to 80 percent, by replacing a randomly located square block of each test image with an unrelated image; see Figure \ref{fig: EYaleB_occlusion} for example. The size of non-overlapping blocks in the PCANet is set to 8$\times$8. We compare with LBP \cite{Ahonen2006} and LBP of the test images being processed by illumination normalization, P-LBP \cite{Tan2010}. We use the NN classifier with the chi-square distance measure.

The experimental results are given in Table \ref{table: ExtendedYaleB}. One can see that the PCANet outperforms the P-LBP for different levels of occlusion. It is also observed that the PCANet is not only illumination-insensitive, but also robust against block occlusion. Under such a single sample per person setting and various difficult lighting conditions, the PCANet surprisingly achieves almost perfect recognition 99.58$\%$, and still sustains 86.49$\%$ accuracy when 60$\%$ pixels of every test image are occluded! The reason could be that each PCA filter can be seen as a detector with the maximum response for patches from a face. In other words, the contribution from occluded patches would somehow be ignored after PCA filtering and are not passed onto the output layer of the PCANet, thereby yielding striking robustness to occlusion.

\begin{table}\centering
\caption{Recognition rates $(\%)$ on Extended Yale B dataset.}
\begin{tabular}{l|c|c|c|c|c}
  \hline
  % after \\: \hline or \cline{col1-col2} \cline{col3-col4} ...
  Percent occluded & 0$\%$ & 20$\%$  & 40$\%$  & 60$\%$  & 80$\%$  \\ \hline \hline
  LBP \cite{Ahonen2006} & 75.76 & 65.66 & 54.92 & 43.22  & 18.06 \\
  P-LBP \cite{Tan2010}    & 96.13 & 91.84 & 84.13 & 70.96 & 41.29 \\ \hline
%  RandNet-1 & 92.05 & 90.74  & 85.14 & 71.51 & 40.03 \\
%  RandNet-2  & 99.03 & 96.76  & 93.60 & 79.46  & 36.87 \\
  PCANet-1   & 97.77 & 96.34 & 93.81 & 84.60 & {\bf  54.38} \\
  PCANet-2   & {\bf 99.58} & {\bf  99.16} & {\bf 96.30} & {\bf 86.49} & 51.73 \\
  \hline
\end{tabular}\label{table: ExtendedYaleB}
\end{table}

\subsubsection{Testing on AR Dataset.}\label{sec: AR}
We further evaluate the ability of the MultiPIE-learned PCANet to cope with real possibly malicious occlusions using AR dataset \cite{Martinez1998}. AR dataset consists of over 4,000 frontal images for 126 subjects. % For each individual, 26 pictures were taken in two separate sessions.
These images include different facial expressions, illumination conditions and disguises. In the experiment, we chose a subset of the data consisting of 50 male subjects and 50 female subjects. The images are cropped with dimension 165$\times$120 and converted to gray scale. For each subject, we select the face with frontal illumination and neural expression in the gallery training, and the rest are all for testing. The size of non-overlapping blocks in the PCANet is set to 8$\times$6. We also compare with LBP \cite{Ahonen2006} and P-LBP \cite{Tan2010}. We use the NN classifier with the chi-square distance measure.

The results are given in Table \ref{tab: AR_dataset}. For test set of illumination variations, the recognition by PCANet is again almost perfect, and for cross-disguise related test sets, the accuracy is more than 95$\%$. The results are consistent with that on MultiPIE and Extended Yale B datasets: PCANet is insensitive to illumination and robust to occlusions. To the best of our knowledge, no single feature with a simple classifier can achieve such performances, even if in extended representation-based classification (ESRC) \cite{Deng2012}! %Gabor features with auxiliary dictionaries could be possible to achieve comparable performance.

\begin{table}\centering
\caption{Recognition rates $(\%)$ on AR dataset.}
\begin{tabular}{l|c|c|c|c}
  \hline
  % after \\: \hline or \cline{col1-col2} \cline{col3-col4} ...
  Test sets & Illum. & Exps.  & Disguise  & Disguise + Illum.  \\ \hline \hline
  LBP \cite{Ahonen2006} & 93.83 & 81.33 & 91.25 & 79.63 \\
  P-LBP \cite{Tan2010}    & 97.50 & 80.33 & 93.00 & 88.58 \\ \hline
%  RandNet-1    & 95.50   &  76.67   &     93.25   &   86.63 \\
%  RandNet-2    &  98.50  &   83.33   &     95.00  &   91.63  \\
  PCANet-1   & 98.00 & {\bf 85.67} & 95.75 & 92.75 \\
  PCANet-2   & {\bf 99.50} & 85.00 & {\bf 97.00} & {\bf 95.00}  \\
  \hline
\end{tabular}\label{tab: AR_dataset}
\end{table}

\subsubsection{Testing on FERET Dataset.}\label{sec: FERET}
We finally apply the MultiPIE-learned PCANet to the popular FERET dataset \cite{Jonathon1998}, which is a standard dataset used for facial recognition system evaluation. FERET contains images of 1,196 different individuals with up to 5 images of each individual captured under different lighting conditions, with non-neural expressions and over the period of three years. The complete dataset is partitioned into disjoint sets: gallery and probe. The probe set is further subdivided into four categories: {\it Fb} with different expression changes; {\it Fc} with different lighting conditions; {\it Dup-I} taken within the period of three to four months; {\it Dup-II} taken at least one and a half year apart. We use the gray-scale images, cropped to image size of 150$\times$90 pixels. The size of non-overlapping blocks in the PCANet is set to {15$\times$15}. To compare fairly with prior methods, the dimension of the PCANet features are reduced to 1000 by a whitening PCA (WPCA),\footnote{The PCA projection directions are weighted by the inverse of their corresponding square-root energies, respectively.} where the projection matrix is learned from the features of gallery samples. The NN classifier with cosine distance is used. Moreover, in addition to PCANet trained from MultiPIE database, we also train PCANet on the FERET generic training set, consisting of 1,002 images of 429 people listed in the FERET standard training CD.

{
The results of the PCANet and other state-of-the-art methods are listed in Table \ref{tab: FERET_dataset}. Surprisingly, both simple MultiPIE-learned PCANet-2 and FERET-learned PCANet-2 (with Trn. CD in a parentheses) achieve the state-of-the-art accuracies 97.25$\%$ and 97.26$\%$ on average, respectively. As the variations in MultiPIE database are much richer than the standard FERET training set, it is nature to see that the MultiPIE-learned PCANet slightly outperforms FERET-learned PCANet. More importantly, PCANet-2 breaks the records in {\it Dup-I} and {\it Dup-II}.
}

{\bf Conclusive remarks on face recognition.} A prominent message drawn from the above experiments in sections \ref{sec: MultiPIE}, \ref{sec: ExtendedYaleB}, \ref{sec: AR}, and \ref{sec: FERET} is that training PCANet from a face dataset can be very effective to capture the abstract representation of new subjects or new datasets. After the PCANet is trained, extracting PCANet-2 feature for one test face only takes 0.3 second in Matlab. We can anticipate that the performance of PCANet could be further improved and moved toward practical use if the PCANet is trained upon a wide and deep dataset that collect sufficiently many inter-class and intra-class variations.

\begin{table}\centering
\caption{Recognition rates $(\%)$ on FERET dataset.}
\begin{tabular}{l|c|c|c|c||c}
  \hline
  % after \\: \hline or \cline{col1-col2} \cline{col3-col4} ...
  Probe sets & {\it Fb} & {\it Fc}  & {\it Dup-I}  &  {\it Dup-II} & {\rm Avg.} \\ \hline \hline
  LBP \cite{Ahonen2006} & 93.00 & 51.00 & 61.00 & 50.00 & 63.75\\
  DMMA \cite{Lu2013}  & 98.10 & 98.50 & 81.60 & 83.20 & 89.60 \\
  P-LBP \cite{Tan2010}    & 98.00 & 98.00 & 90.00 & 85.00 & 92.75 \\
  POEM \cite{Vu2012}    & 99.60 & 99.50 & 88.80 & 85.00 & 93.20 \\
  G-LQP \cite{Hussain2012} & {\bf 99.90} & {\bf  100} & 93.20 & 91.00 & 96.03 \\
  LGBP-LGXP \cite{Xie2010} & 99.00 & 99.00 & 94.00 & 93.00 & 96.25 \\
  sPOEM+POD \cite{Vu2013}   &99.70  & {\bf 100} & 94.90   & 94.00 & 97.15  \\
  GOM \cite{Chai2014} & 99.90 & {\bf 100} &  95.70  & 93.10 & 97.18 \\ \hline
  PCANet-1 (Trn. CD) & 99.33  &  99.48  &  88.92  &  84.19 & 92.98 \\
  PCANet-2 (Trn. CD) &  99.67  &  99.48  & {\bf  95.84}  &  {\bf 94.02} & 97.25 \\
%  RandNet-1 & 98.24  &  94.33   & 85.60  &  80.34 & 89.63 \\
%  RandNet-2   &    98.41  &  99.48  &  93.49  &  91.03 & 95.60 \\
  PCANet-1   & 99.50  &  98.97 &   89.89  &  86.75 &  93.78 \\
  PCANet-2   & 99.58 & {\bf 100}    &  95.43 & {\bf  94.02} & {\bf 97.26}  \\
  \hline
\end{tabular}\label{tab: FERET_dataset}
\end{table}

\subsection{Face Verification on LFW Dataset}
Besides tests with laboratory face datasets, we also evaluate the PCANet on the LFW dataset \cite{Huang2007} for unconstrained face verification. LFW contains 13,233 face images of 5,749 different individuals, collected from the web with large variations in pose, expression, illumination, clothing, hairstyles, etc. We consider ``unsupervised setting'', which is the best choice for evaluating the learned features, for it does not depend on metric learning and discriminative model learning. The aligned version of the faces, namely LFW-a, provided by Wolf {\it et al.} \cite{Wolf2011} is used, and the face images were cropped into $150 \times 80$ pixel dimensions. We follow the standard evaluation protocal, which splits the View 2 dataset into 10 subsets with each subset containing 300 intra-class pairs and 300 inter-class pairs. We perform 10-fold cross validation using the 10 subsets of pairs in View 2. In PCANet, the filter size, the number of filters, and the (non-overlapping) block size are set to $k_1 = k_2 = 7$, $L_1 = L_2 = 8$, and 15$\times$13, respectively. The performances are measured by averaging the 10-fold cross validation. We project the PCANet features onto 400 and 3,200 dimensions using WPCA for PCANet-1 and PCANet-2, respectively, and use NN classifier with cosine distance.

{
Table \ref{table: LFW} tabulates the results.\footnote{For fair comparison, we only report results of single descriptor. The best known LFW result under unsupervised setting is 88.57\% \cite{Barkan2013}, which is inferred from four different descriptors. } Note that PCANet followed by sqrt in a parentheses represents the PCANet feature taking square-root operation. One can see that the square-root PCANet outperforms PCANet, and this performance boost from square-root operation has also been observed in other features for this dataset \cite{Barkan2013}. Moreover, the square-root PCANet-2 that achieves 86.28$\%$ accuracy is quite competitive to the current state-of-the-art methods.
%It is on par with all the highly hand-crafted features for this dataset and it is within less than two percents of the best features of \cite{Hussain2012} on this dataset, but arguably better than \cite{Hussain2012} on the FERET dataset.
This shows that the proposed PCANet is also effective in learning invariant features for face images captured in less controlled conditions.

In preparation of this paper, we are aware of two concurrent works \cite{Taigman2014, Fan2014} that employ ConvNet for LFW face verification. While both works achieve very impressive results on LFW, their experimental setting differs from ours largely. These two works require some outside database to train the ConvNet and the face images have to be more precisely aligned; e.g., \cite{Taigman2014} uses 3-dimensional model for face alignment and \cite{Fan2014} extracts multi-scale features based on detected landmark positions. On the contrary, we only trained PCANet based on LFW-a  \cite{Wolf2011}, an aligned version of LFW images using the commercial alignment system of face.com. }

\begin{table}[htbp]\centering
\caption{Comparison of verification rates $(\%)$ on LFW under unsupervised setting.}
\begin{tabular}{l|c}
  \hline
  % after \\: \hline or \cline{col1-col2} \cline{col3-col4} ...
  Methods           & Accuracy \\  \hline \hline
  POEM \cite{Vu2012}  & 82.70$\pm$0.59 \\
  High-dim. LBP \cite{Chen2013}  & 84.08 \\
  High-dim. LE \cite{Chen2013}  & 84.58 \\
  SFRD \cite{Cui2013}  & 84.81  \\
  I-LQP \cite{Hussain2012} & 86.20$\pm$0.46 \\
  OCLBP \cite{Barkan2013} & {\bf 86.66$\pm$0.30} \\  \hline
%  RandNet-1 & 78.57 $\pm$ 1.66 \\
%  RandNet-2 & 83.22 $\pm$ 1.22 \\
  PCANet-1  & 81.18 $\pm$ 1.99  \\
  PCANet-1 ({\rm sqrt})  & 82.55 $\pm$ 1.48  \\
  PCANet-2  & 85.20 $\pm$ 1.46  \\
  PCANet-2 ({\rm sqrt}) & 86.28 $\pm$ 1.14  \\
  \hline
\end{tabular}\label{table: LFW}
\end{table}

\subsection{Digit Recognition on MNIST Datasets}
We now move forward to test the proposed PCANet, along with RandNet and LDANet, on MNIST \cite{LeCun1998} and MNIST variations \cite{Larochelle2007}, a widely-used benchmark for testing hierarchical representations. There are 9 classification tasks in total, as listed in Table \ref{table: mnist_details}. All the images are of size $28\times 28$. In the following, we use {\em MNIST basic} as the dataset to investigate the influence of the number of filters or different block overlap ratios for RandNet, PCANet and LDANet, and then compare with other state-of-the-art methods on all the MNIST datasets.

\subsubsection{Impact of the number of filters}
\begin{figure}[t]
\centering
\subfigure[tight][]{
\resizebox{0.45\linewidth}{!}{\includegraphics{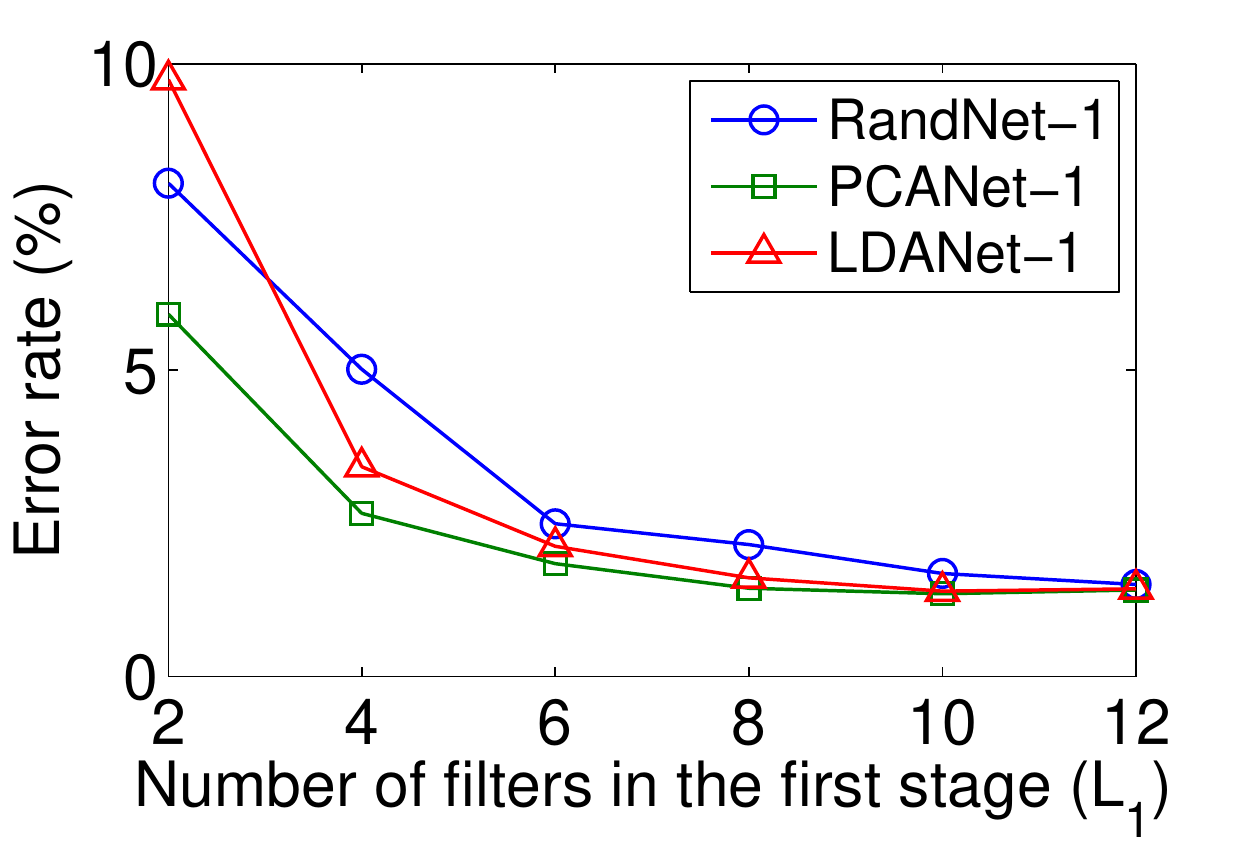}}\label{fig:
Nets_K1}}\hspace{0cm} \subfigure[tight][]{
\resizebox{0.45\linewidth}{!}{\includegraphics{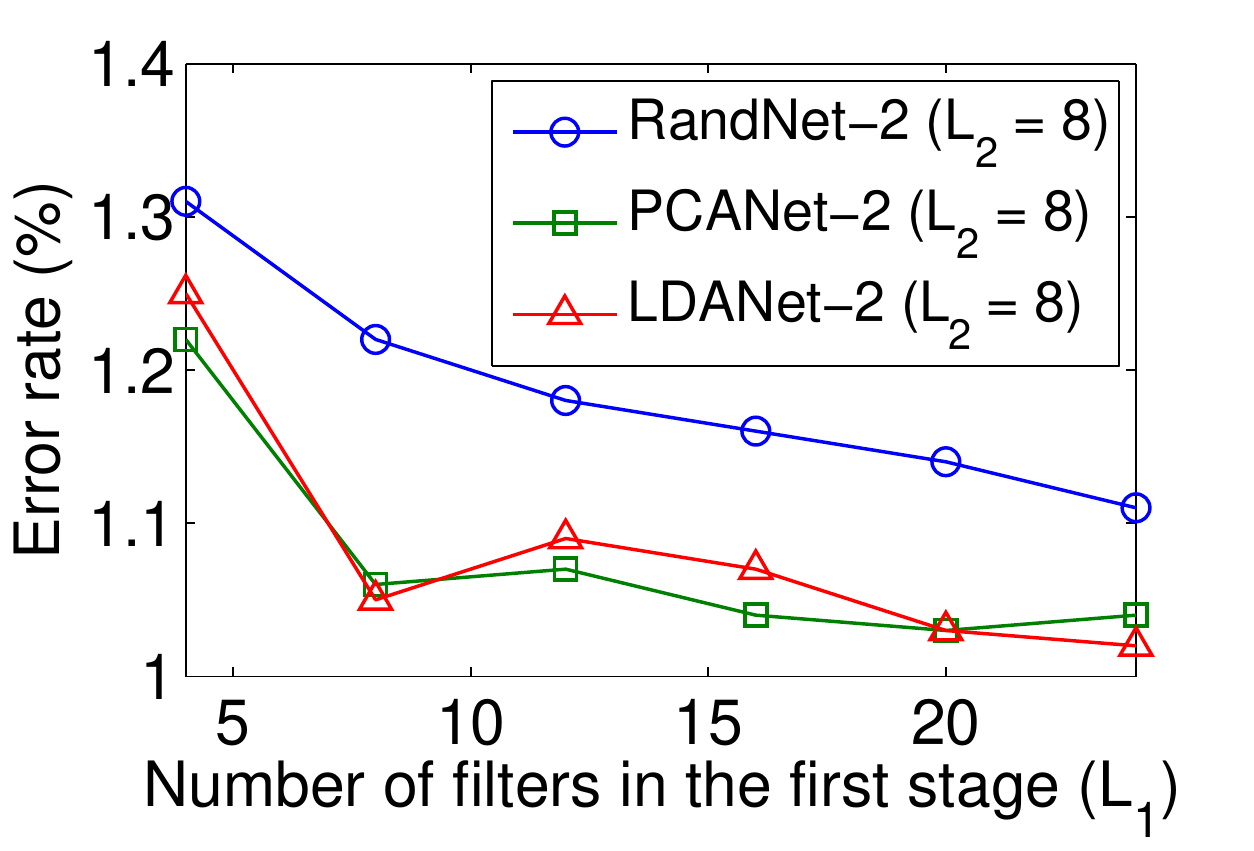}}\label{fig:
Nets_K1_8}}
\caption{Error rate of PCANet on MNIST basic test set for varying number of filters in the first stage. (a) PCANet-1; (b) PCANet-2 with $L_2 = 8$.}\label{fig: mnist_basic_num_filters}
\end{figure}

We vary the number of filters in the first stage $L_1$ from $2$ to $12$ for one-stage networks. Regarding two-stage networks, we set $L_2=8$ and change $L_1$ from $4$ to $24$. The filter size of the networks is $k_1 = k_2 = 7$, block size is 7$\times$7, and the overlapping region between blocks is half of the block size. The results are shown in Figure \ref{fig: mnist_basic_num_filters}. The results are consistent with that for MultiPIE face database in Figure \ref{fig: MultiPIE_K1}; PCANet outperforms RandNet and LDANet for almost all the cases.

\subsubsection{Impact of the block overlap ratio}
The number of filters is fixed to $L_1 = L_2 = 8$, and the filter size is again $k_1 = k_2 = 7$ and block size is 7$\times$7. We only vary the block overlap ratio (BOR) from $0.1$ to $0.7$. Table \ref{table: MNIST_basic_blockratio} tabulates the results of RandNet-2, PCANet-2, and LDANet-2. Clearly, PCANet-2 and LDANet-2 achieve their minimum error rates for BOR equal to 0.5 and 0.6, respectively, and PCANet-2 performs the best for all conditions.

\begin{table}\centering
\caption{Error rates $(\%)$ of PCANet-2 on {\em basic} dataset for varying block overlap ratios (BORs). }
\vspace{0.1cm}\begin{tabular}{c|c|c|c|c|c|c|c}
  \hline
  % after \\: \hline or \cline{col1-col2} \cline{col3-col4} ...
  BOR  & 0.1 & 0.2 & 0.3 & 0.4 & 0.5 & 0.6 & 0.7   \\ \hline \hline
  RandNet-2  & 1.31  & 1.35 & 1.23  & 1.34  & 1.18 & 1.14 & 1.24 \\
  PCANet-2   & {\bf 1.12} & {\bf 1.12} & {\bf 1.07} & 1.06 & 1.06 & {\bf 1.02}  & {\bf 1.05}  \\
  LDANet-2   & 1.14  &  1.14  &  1.11  &   {\bf 1.05}  &  {\bf 1.05}  &  1.05 &   1.06 \\
  \hline
\end{tabular}\label{table: MNIST_basic_blockratio}
\end{table}

\subsubsection{Comparison with state of the arts}
We compare RandNet, PCANet, and LDANet with ConvNet \cite{Jarrett2009}, 2-stage ScatNet (ScatNet-2) \cite{Bruna2013}, and other existing methods. In ScatNet, the number of scales and the number of orientations are set to 3 and 8, respectively. Regarding the parameters of PCANet, we set the filter size $k_1 = k_2 = 7$, the number of PCA filters $L_1 = L_2 = 8$; the block size is tuned by a cross-validation for {\it MNIST}, and the validation sets for MNIST variations\footnote{Using either cross-validation or validation set, the optimal block size is obtained as 7$\times$7 for {\it MNIST}, {\it basic}, {\it rec-img}, 4$\times$4 for {\it rot}, {\it bg-img}, {\it bg-rnd}, {\it bg-img-rot}, 14$\times$14 for {\it rec}, and 28$\times$28 for {\it convex}.}. The overlapping region between blocks is half of the block size. Unless otherwise specified, we use linear SVM classifier for ScatNet and RandNet, PCANet and LDANet for the 9 classification tasks.

The testing error rates of the various methods on {\it MNIST} are shown in Table \ref{table: mnist_standard}. For fair comparison, we do not include the results of methods using augmented training samples with distortions or other information, for that the best known result is 0.23$\%$ \cite{Ciresan2012}. We see that RandNet-2, PCANet-2, and LDANet-2 are comparable with the state-of-the-art methods on this standard MNIST task.
However, as {\em MNIST} has many training data, all methods perform very well and very close -- the difference is not so statistically meaningful.

Accordingly, we also report results of different methods on MNIST variations in Table \ref{table: mnist}. To the best of our knowledge, the PCANet-2 achieves the state-of-the-art results for four out of the eight remaining tasks: {\it basic}, {\it bg-img}, {\it bg-img-rot}, and {\it convex}. Especially for {\it bg-img}, the error rate reduces from 12.25$\%$ \cite{Sohn2013} to 10.95$\%$.

Table \ref{table: mnist} also shows the result of PCANet-1 with $L_1L_2$ filters of size $(k_1 + k_2 -1) \times (k_1 + k_2 -1)$. The PCANet-1 with such a parameter setting is to mimic the reported PCANet-2 in a single-stage structure. PCANet-2 still outperforms this PCANet-1 alternative.

\begin{figure}[t]
\centering
\resizebox{0.9\linewidth}{!}{\includegraphics{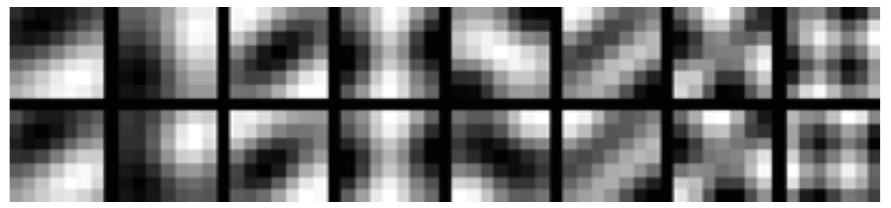}}
\caption{The PCANet filters learned on MNIST dataset. Top row: the first stage. Bottom row: the second stage. The filter size $k_1 = k_2 = 7$ are set in RandNet, PCANet, and LDANet unless specified otherwise.}\label{fig: MNIST_filter}
\end{figure}

\begin{figure*}[t]
\centering
\resizebox{0.8\linewidth}{!}{\includegraphics{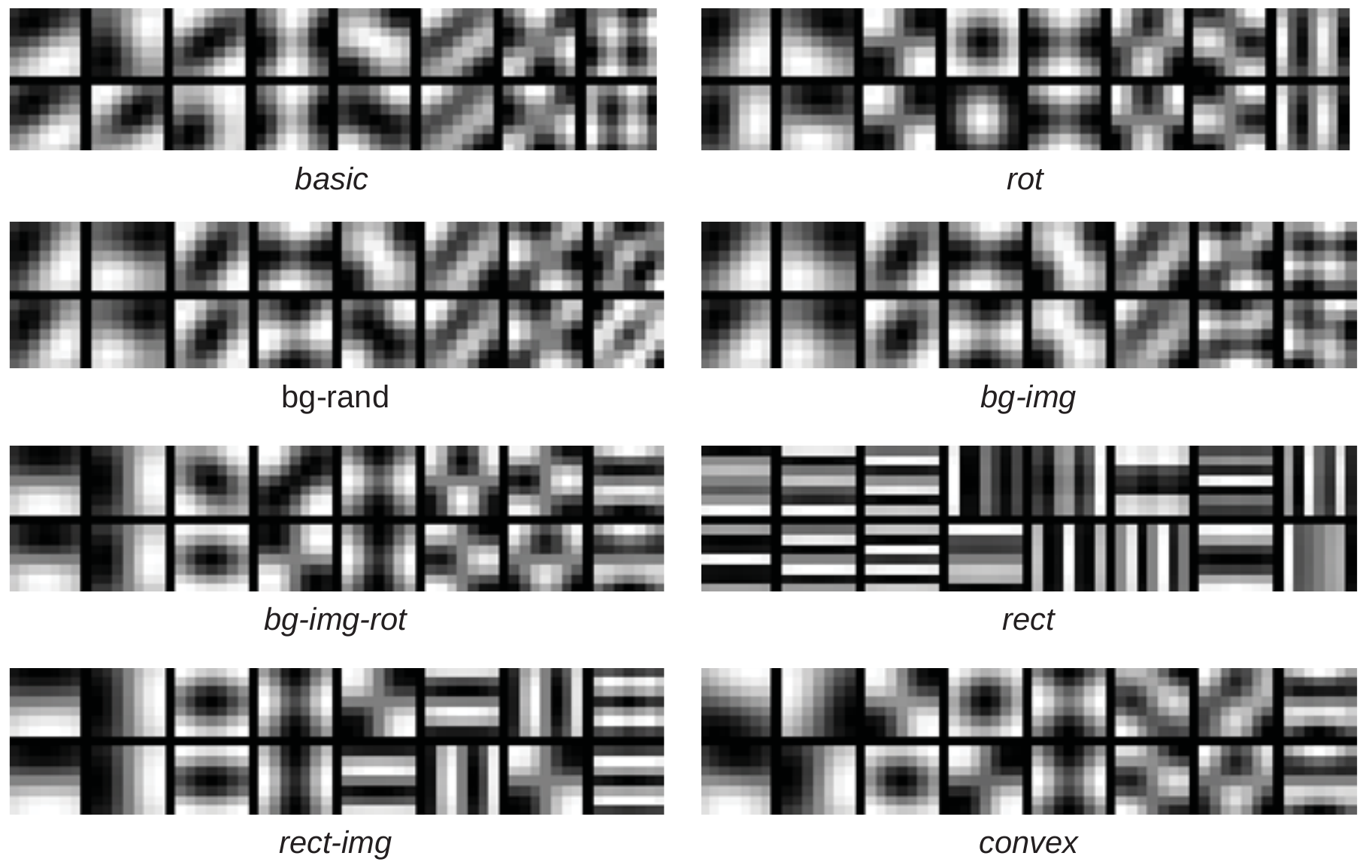}}
\caption{The PCANet filters learned on various MNIST datasets. For each dataset, the top row shows the filters of the first stage; the bottom row shows the filters of the second stage.} \label{fig: mnist_filters}
\end{figure*}

Furthermore, we also draw the learned PCANet filters in Figure \ref{fig: MNIST_filter} and Figure \ref{fig: mnist_filters}. An intriguing pattern is observed in the filters of {\em rect} and {\em rect-img} datasets. For {\em rect}, we can see both horizontal and vertical stripes, for these patterns attempt to capture the edges of the rectangles. When there is some image background in {\em rect-img}, several filters become low-pass, in order to secure the responses from background images.

\begin{table*}[htbp]\caption{Details of the 9 classification tasks on MNIST and MNIST variations.}\label{table: mnist_details}
\centering
%\resizebox{0.7\linewidth}{!}{
\begin{tabular}{l|l|c|l}
   \hline
   % after \\: \hline or \cline{col1-col2} \cline{col3-col4} ...
   {Data Sets}   & { Description} & { Num. of classes} & { Train-Valid-Test} \\ \hline \hline
   {\it MNIST}       & {\rm Standard MNIST} & 10 & 60000-0-10000 \\
   {\it basic}       & {\rm Smaller subset of MNIST} & 10 & 10000-2000-50000 \\
   {\it rot}         & {\rm MNIST with rotation} & 10 & 10000-2000-50000 \\
   {\it bg-rand}     & {\rm MNIST with noise background} & 10 & 10000-2000-50000 \\
   {\it bg-img}      & {\rm MNIST with image background} & 10 & 10000-2000-50000 \\
   {\it bg-img-rot}  & {\rm MNIST with rotation and image background} & 10 & 10000-2000-50000 \\
   {\it rect}        & {\rm Discriminate between tall and wide rectangles} & 2 & 1000-200-50000 \\
   {\it rect-img}    & {\rm Dataset {\it rect} with image background} & 2 & 10000-2000-50000 \\
   {\it convex}      & {\rm Discriminate between convex and concave shape} & 2 & 6000-2000-50000 \\
   \hline
\end{tabular}%}
\end{table*}

\begin{table}[htbp]\centering
\caption{Comparison of error rates $(\%)$ of the methods on {\em MNIST}, excluding methods that augment the training data. The filter size $k_1 = k_2 = 7$ are set in RandNet, PCANet, and LDANet unless specified otherwise.}
\begin{tabular}{l|c}
  \hline
  % after \\: \hline or \cline{col1-col2} \cline{col3-col4} ...
  Methods           & {\it MNIST} \\  \hline \hline
  HSC \cite{Yu2011}  &    0.77  \\
  K-NN-SCM \cite{Belongie2002} &  0.63 \\
  K-NN-IDM \cite{Keysers2007} & 0.54 \\
  CDBN \cite{Lee2009}           & 0.82  \\
  ConvNet \cite{Jarrett2009}          & 0.53  \\
  Stochastic pooling ConvNet \cite{Zeiler2013} & 0.47   \\
  Conv. Maxout $+$ Dropout \cite{Goodfellow2013} & 0.45   \\
  ScatNet-2 (SVM$_{rbf}$) \cite{Bruna2013}      & {\bf 0.43}  \\ \hline
  RandNet-1 &  1.32 \\
  RandNet-2 &  0.63 \\
  PCANet-1    & 0.94  \\
  PCANet-2    & 0.66  \\
  LDANet-1   & 0.98 \\
  LDANet-2  &  0.62 \\ \hline
  PCANet-1 ($k_1 = 13$) & 0.62 \\
  %PCANet-2 (SVM$_{rbf}$)   & 0.58  \\
  \hline
\end{tabular}\label{table: mnist_standard}
\end{table}

\begin{table*}\centering
\caption{Comparison of testing error rates $(\%)$ of the various methods on MNIST variations.}
\vspace{0.1cm}\centering\begin{tabular}{l|c|c|c|c|c|c|c|c}
  \hline
  % after \\: \hline or \cline{col1-col2} \cline{col3-col4} ...
  Methods        & {\it basic} & {\it rot}             & {\it bg-rand} & {\it bg-img} & {\it bg-img-rot} & {\it rect} & {\it rect-img} & {\it convex} \\ \hline \hline
  CAE-2 \cite{Rifai2011}        &  2.48  & 9.66         & 10.90       &  15.50     & 45.23       & 1.21 & 21.54 & - \\
  TIRBM \cite{Sohn2012}         &  -     & {\bf 4.20}   &   -         &    -       & 35.50       &   -      &    -   &  - \\
  PGBM $+$ DN-1  \cite{Sohn2013}&   -    &      -       & {\bf 6.08}  &   12.25    &  36.76      & -   &   - &  - \\
  ScatNet-2 \cite{Bruna2013}    & 1.27   & 7.48         &  12.30      & 18.40      & 50.48       & {\bf 0.01} & {\bf 8.02} & 6.50 \\ \hline
  RandNet-1                     &   1.86 & 14.25        &    18.81    & 15.97      &  51.82      &  0.21  &  15.94 &   6.78 \\
  RandNet-2                     & 1.25   & 8.47         &    13.47    & 11.65      &  43.69      &  0.09  &  17.00  &  5.45   \\
  PCANet-1                      & 1.44   & 10.55        &  6.77       & 11.11      & 42.03       & 0.15  & 25.55  & 5.93 \\
  PCANet-2                      & 1.06   & 7.37         &  6.19       & {\bf 10.95}& {\bf 35.48} & 0.24 & 14.08 & {\bf 4.36} \\
  LDANet-1                      & 1.61   &   11.40      &     7.16    &  13.03     &   43.86     &    0.15 &   23.63 &    6.89 \\
  LDANet-2                      &{\bf 1.05} & 7.52      &     6.81    &  12.42     &   38.54     &    0.14 &   16.20 &    7.22 \\ \hline
  PCANet-1 ($k_1 = 13$)         & 1.21   & 8.30         & 6.88        &  11.97     &   39.06    &    0.03  &  13.94  &  6.75  \\
  \hline
\end{tabular}\label{table: mnist}
\end{table*}

\subsection{Texture Classification on CUReT Dataset}
The CUReT texture dataset contains 61 classes of image textures. Each texture class has images of the same material with different pose and illumination conditions. Other than the above variations, specularities, shadowing and surface normal variations also make this classification challenging. In this experiment, a subset of the dataset with azimuthal viewing angle less than 60 degrees is selected, thereby yielding 92 images in each class. A central $200\times 200$ region is cropped from each of the selected images. The dataset is randomly split into a training and a testing set, with $46$ training images for each class, as in \cite{Varma2009}. The PCANet is trained with filter size $k_1 = k_2 = 5$, the number of filters $L_1 = L_2 = 8$, and block size 50$\times$50. We use linear SVM classifier. The testing error rates averaged over 10 different random splits are shown in Table \ref{table: CUReT}. We see that the PCANet-1 outperforms ScatNet-1, but the improvement from PCANet-1 to PCANet-2 is not as large as that of ScatNet. Note that ScatNet-2 followed by a PCA-based classifier gives the best result \cite{Bruna2013}.

\begin{figure}[t]
\centering
\resizebox{0.8\linewidth}{!}{\includegraphics{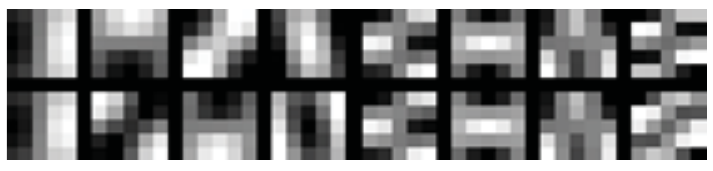}}
\caption{The PCANet filters learned on CUReT database. Top row: the first stage. Bottom row: the second stage. }\label{fig: CUReT_Filters}
\end{figure}

\begin{table}[tbp]\centering
\caption{Comparison of error rates $(\%)$ on CUReT.}
\begin{tabular}{l|c}
  \hline
  % after \\: \hline or \cline{col1-col2} \cline{col3-col4} ...
  Methods           & Error rates \\  \hline \hline
  Textons \cite{Hayman2004} & 1.50 \\
  BIF \cite{Crosier2010}  & 1.40 \\
  Histogram \cite{Broadhurst2006} & 1.00 \\
  ScatNet-1 (PCA) \cite{Bruna2013} & 0.50 \\
  ScatNet-2 (PCA) \cite{Bruna2013} & {\bf 0.20} \\ \hline
  RandNet-1  & 0.61 \\
  RandNet-2 & 0.46 \\
  PCANet-1    & 0.45  \\
  PCANet-2    & 0.39 \\
  LDANet-1    & 0.69  \\
  LDANet-2    & 0.54 \\
  \hline
\end{tabular}\label{table: CUReT}
\end{table}

\subsection{Object Recognition on CIFAR10}
We finally evaluate the performance of PCANet on CIFAR10 database for object recognition. CIFAR10 is a set of natural RGB images of 32$\times$32 pixels. It contains 10 classes with 50000 training samples and 10000 test samples. Images in CIFAR10 vary significantly not only in object position and object scale within each class, but also in colors and textures of these objects.

\begin{figure}[t]
\centering
\resizebox{0.9\linewidth}{!}{\includegraphics{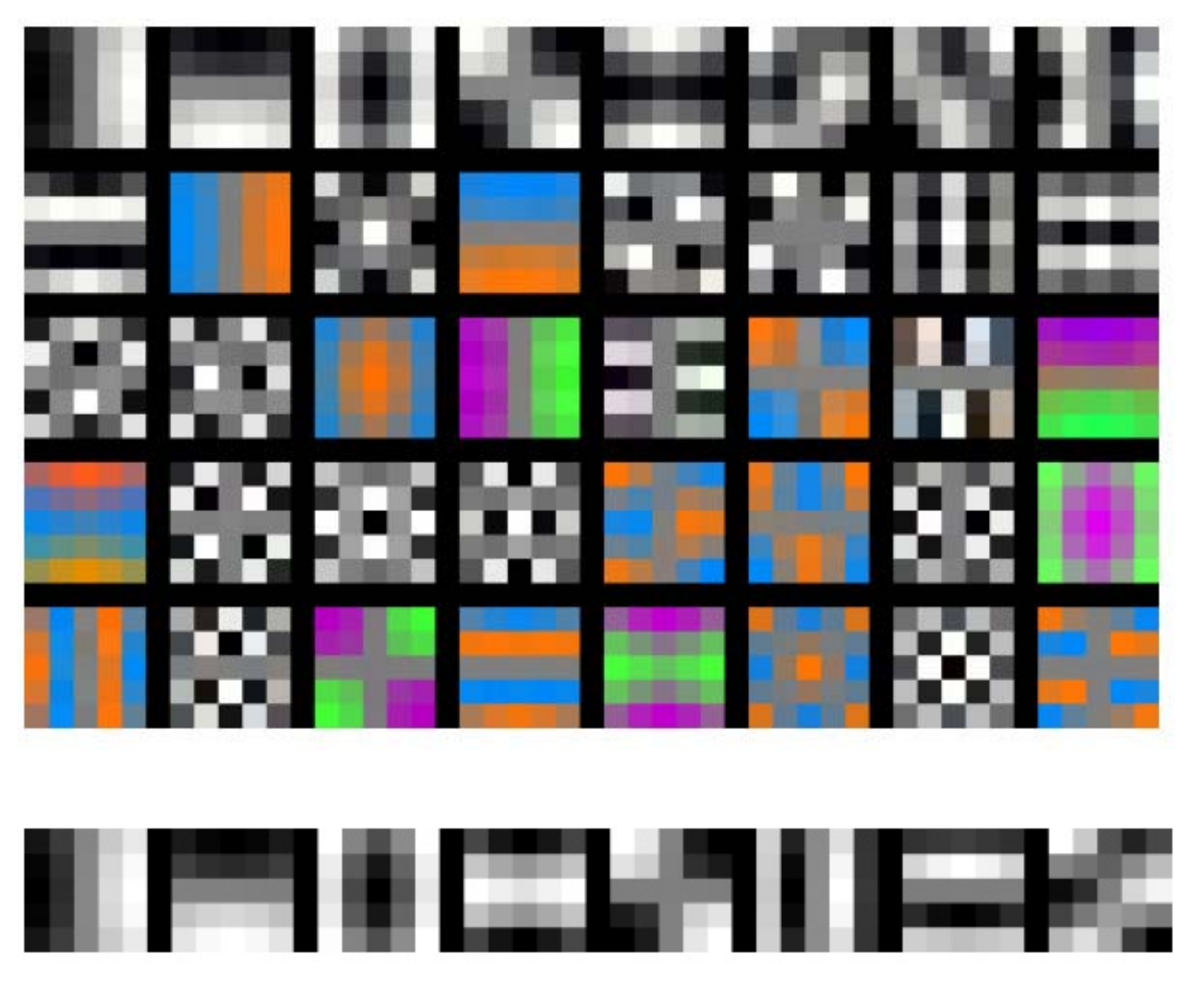}}
\caption{The PCANet filters learned on Cifar10 database. Top: the first stage. Bottom: the second stage. }\label{fig: cifar10_filters}
\end{figure}

The motivation here is to explore the limitation of such a simple PCANet on a relatively complex database, in comparison to the databases of faces, digits, and textures we have experimented with, which could somehow be roughly aligned or prepared. To begin with, we extend PCA filter learning so as to accommodate the RGB images in object databases. In the same spirit of constructing the data matrix $\bm{X}$ in \eqref{eq: datamatrix_1}, we gather the same individual matrix for RGB channels of the images, denoted by $\bm{X}_{\rm r}, \bm{X}_{\rm g}, \bm{X}_{\rm b} \in \mathbb{R}^{k_1k_2 \times Nmn}$, respectively. Following the key steps in Section \ref{sec: The first stage}, the multichannel PCA filters can be easily verified as
\begin{equation}\label{eq: multichannel_PCAfilter1}
\bm{W}^{\rm r,g,b}_l \doteq {\rm mat}_{k_1,k_2,3}(\bm{ q}_l(\widetilde{\bm{X}}\widetilde{\bm{X}}^T))\in\mathbb{R}^{k_1 \times k_2 \times 3},
\end{equation}where $\widetilde{\bm{X}} = [\bm{X}_{\rm r}^T, \bm{X}_{\rm g}^T, \bm{X}_{\rm b}^T]^T$ and ${\rm mat}_{k_1,k_2,3}(\bm{ v})$ is a function that maps $\bm{ v}\in \mathbb{R}^{3k_1k_2 }$ to a tensor $\bm{ W}\in \mathbb{R}^{k_1 \times k_2 \times 3}$. An example of the learned multichannel PCA filters is demonstrated in Figure \ref{fig: cifar10_filters}. In addition to the modification above, we also connect spatial pyramid pooling (SPP) \cite{Grauman2005, Lazebnik2006, He2014} to the output layer of PCANet, with the aim of extracting information invariant to large poses and complex backgrounds, usually seen in object databases. The SPP essentially helps object recognition, but finds no significant improvement in the previous experiments on faces, digits and textures.

We use linear SVM classifier in the experiments. In the first experiment, we train PCANet on CIFAR10 with filter size $k_1 = k_2 = 5$, the number of filters $L_1 = 40,~L_2 = 8$, and block size equal to $8\times 8$. Also, we set the overlapping region between blocks to half of the block size, and connected SPP to the output layer of PCANet; i.e., the maximum response in each bin of block histograms is pooled in a pyramid of 4$\times$4, 2$\times$2, and 1$\times$1 subregions. This yields the $21$ pooled histogram feature of dimension $L_1 2^{L_2}$. The dimension of each pooled feature is reduced to 1280 by PCA.

In the second experiment, we concatenate PCANet features learned with different filter size $k_1 = k_2 = 3$ and $k_1 = k_2 = 5$. All the processes and model parameters are fixed identical to the single descriptor mentioned in last paragraph, except $L_1 = 12$ and $L_1 = 28$ set for filter size equal to $3$ and $5$, respectively. This is to ensure that the combined features are of the same dimension with the single descriptor, for fairness.

The results are shown in Table \ref{table: cifar10}. PCANet-2 achieves accuracy 77.14$\%$ and gains 1.5$\%$ improvement when combining two features learned with different filter sizes (marked with combined in a parenthesis). While PCANet-2 has around 11$\%$ accuracy degradation in comparison to state-of-the-art method (with no data augmentation), the performance of the fully unsupervised and extremely simple PCANet-2 shown here is still encouraging.

\begin{table}[tbp]\centering
\caption{Comparison of accuracy $(\%)$ of the methods on CIFAR10 with no data augmentation.}
\begin{tabular}{l|c}
  \hline
  % after \\: \hline or \cline{col1-col2} \cline{col3-col4} ...
  Methods         &  Accuracy   \\  \hline \hline
  Tiled CNN \cite{Le2010} & 73.10 \\
  Improved LCC \cite{Yu2010} & 74.50 \\
  KDES-A \cite{Bo2010} & 76.00 \\
  K-means (Triangle, 4000 features) \cite{Coates2010} & 79.60 \\
  Cuda-convnet2 \cite{Krizhevsky2014} & 82.00 \\
  Stochastic pooling ConvNet \cite{Zeiler2013} & 84.87   \\
  CNN $+$ Spearmint \cite{Snoek2012} & 85.02 \\
  Conv. Maxout $+$ Dropout \cite{Goodfellow2013} & 88.32   \\
  NIN \cite{Lin2014}  & 89.59      \\  \hline
  PCANet-2  & 77.14  \\
  PCANet-2 (combined) & 78.67 \\
  \hline
\end{tabular}\label{table: cifar10}
\end{table}

\section{Conclusion}
In this paper, we have proposed arguably the simplest unsupervised convolutional deep learning network--- PCANet. The network processes input images by cascaded PCA, binary hashing, and block histograms. Like the most ConvNet models, the network parameters such as the number of layers, the filter size, and the number of filters have to be given to PCANet. Once the parameters are fixed, training PCANet is extremely simple and efficient, for the filter learning in PCANet does not involve regularized parameters and does not require numerical optimization solver. Moreover, building the PCANet comprises only a cascaded linear map, followed by a nonlinear output stage. Such a simplicity offers an alternative and yet refreshing perspective to convolutional deep learning networks, and could further facilitate mathematical analysis and justification of its effectiveness.

A couple of simple extensions of PCANet; that is, RandNet and LDANet, have been introduced and tested together with PCANet on many image classification tasks, including face, hand-written digit, texture, and object. Extensive experimental results have consistently shown
that the PCANet outperforms RandNet and LDANet, and is generally on par with ScatNet and variations of ConvNet. %The reason that ConvNet performs poorly on face recognition tasks is that the discriminatively trained ConvNets might overfit the arguably limited face gallery images.
Furthermore, the performance of PCANet is closely comparable and often better than highly engineered hand-crafted features (such as LBP and LQP). In tasks such as face recognition, PCANet also demonstrates remarkable robustness to corruption and ability to transfer to new datasets.

The experiments also convey that as long as the images in databases are somehow well prepared; i.e., images are roughly aligned and do not exhibit diverse scales or poses, PCANet is able to eliminate the image variability and gives reasonably competitive accuracy. In challenging image databases such as Pascal and ImageNet, PCANet might not be sufficient to handle the variability, given its extremely simple structure and unsupervised learning method. %\footnote{On the contrary, ConvNet and its variations perform quite well in these complex databases, but seems not straightforward to be extended for face recognition tasks.}
An intriguing research direction will then be how to construct a more complicated (say more sophisticated filters possibly with discriminative learning) or deeper (more number of stages) PCANet that could accommodate the aforementioned issues. Some preprocessing of pose alignment and scale normalization might be needed for good performance guarantee. The current bottleneck that keeps PCANet from growing deeper (e.g., more than two stages) is that the dimension of the resulted feature would increase exponentially with the number of stages. This fortunately seems able to be fixed by replacing the 2-dimensional convolution filters with tensor-like filters as in \eqref{eq: multichannel_PCAfilter1}, and it will be our future study. Furthermore, we will also leave as future work to augment PCANet with a simple, scalable baseline classifier, readily applicable to much larger scale datasets or problems.

Regardless, extensive experiments given in this paper sufficiently conclude two facts: 1) the PCANet is a very simple deep learning network, effectively extracting useful information for classification of faces, digits, and texture images; 2) the PCANet can be a valuable baseline for studying advanced deep learning architectures for large-scale image classification tasks.

%{\footnotesize
\bibliographystyle{IEEEtran}
\bibliography{ref_note}

% Generated by IEEEtran.bst, version: 1.13 (2008/09/30)
\begin{thebibliography}{10}
\providecommand{\url}[1]{#1}
\csname url@samestyle\endcsname
\providecommand{\newblock}{\relax}
\providecommand{\bibinfo}[2]{#2}
\providecommand{\BIBentrySTDinterwordspacing}{\spaceskip=0pt\relax}
\providecommand{\BIBentryALTinterwordstretchfactor}{4}
\providecommand{\BIBentryALTinterwordspacing}{\spaceskip=\fontdimen2\font plus
\BIBentryALTinterwordstretchfactor\fontdimen3\font minus
  \fontdimen4\font\relax}
\providecommand{\BIBforeignlanguage}[2]{{%
\expandafter\ifx\csname l@#1\endcsname\relax
\typeout{** WARNING: IEEEtran.bst: No hyphenation pattern has been}%
\typeout{** loaded for the language `#1'. Using the pattern for}%
\typeout{** the default language instead.}%
\else
\language=\csname l@#1\endcsname
\fi
#2}}
\providecommand{\BIBdecl}{\relax}
\BIBdecl

\bibitem{Hinton2006}
G.~Hinton, S.~Osindero, and Y.-W. Teh, ``A fast learning algorithm for deep
  belief nets,'' \emph{Neural Computation}, vol.~18, no.~7, pp. 1527--1554,
  2006.

\bibitem{Bengio2013}
Y.~Bengio, A.~Courville, and P.~Vincent, ``Representation learning: a review
  and new perspectives,'' \emph{IEEE TPAMI}, vol.~35, no.~8, pp. 1798--1828,
  2013.

\bibitem{Goodfellow2013}
I.~J. Goodfellow, D.~Warde-Farley, M.~Mirza, A.~Courville, and Y.~Bengio,
  ``Maxout networks,'' in \emph{ICML}, 2013.

\bibitem{LeCun1998}
Y.~Lecun, L.~Bottou, Y.~Bengio, and P.~Haffner, ``Gradient-based learning
  applied to document recognition,'' \emph{Proceedings of the IEEE}, vol.~86,
  no.~11, pp. 2278--2324, 1998.

\bibitem{Jarrett2009}
K.~Jarrett, K.~Kavukcuoglu, M.~Ranzato, and Y.~LeCun, ``What is the best
  multi-stage architecture for object recognition,'' in \emph{ICCV}, 2009.

\bibitem{Bruna2013}
J.~Bruna and S.~Mallat, ``Invariant scattering convolution networks,''
  \emph{IEEE TPAMI}, vol.~35, no.~8, pp. 1872--1886, 2013.

\bibitem{Lee2009}
H.~Lee, R.~Grosse, R.~Rananth, and A.~Ng, ``Convolutional deep belief networks
  for scalable unsupervised learnig of hierachical representation,'' in
  \emph{ICML}, 2009.

\bibitem{Krizhevsky2012}
A.~Krizhevsky, I.~Sutskever, and G.~Hinton, ``Imagenet classification with deep
  convolutional neural network,'' in \emph{NIPS}, 2012.

\bibitem{Kayukcuoglu2010}
K.~Kavukcuoglu, P.~Sermanet, Y.~Boureau, K.~Gregor, M.~Mathieu, and Y.~LeCun,
  ``Learning convolutional feature hierarchies for visual recognition,'' in
  \emph{NIPS}, 2010.

\bibitem{Sifre2013}
L.~Sifre and S.~Mallat, ``Rotation, scaling and deformation invariant
  scattering for texture discrimination,'' in \emph{CVPR}, 2013.

\bibitem{Burges2003}
C.~J.~C. Burges, J.~C. Platt, and S.~Jana, ``Distortion discriminant analysis
  for audio fingerprinting,'' \emph{IEEE TSAP}, vol.~11, no.~3, pp. 165--174,
  2003.

\bibitem{Lowe2004}
D.~G. Lowe, ``Distinctive image features from scale-invariant keypoints,''
  \emph{International Journal of Computer Vision}, vol.~60, no.~2, pp. 91--110.

\bibitem{Dalal2005}
N.~Dalal and B.~Triggs, ``Histograms of oriented gradients for human
  detection,'' in \emph{IEEE CVPR}, 200.

\bibitem{Fei2005}
L.~Fei-Fei and P.~Perona, ``A {B}ayesian hierarchical model for learning
  natural scene categories,'' in \emph{IEEE CVPR}, 2005.

\bibitem{Liu2002}
C.~Liu and H.~Wechsler, ``Gabor feature based classification using the enhanced
  fisher linear discriminant model for face recognition,'' \emph{IEEE TIP},
  vol.~11, no.~4, pp. 467--476, 2002.

\bibitem{Yu2001}
H.~Yu and J.~Yang, ``A direct {LDA} algorithm for high-dimensional data--- with
  application to face recognition,'' \emph{Pattern Recognition}, vol.~34,
  no.~10, pp. 2067--2069, 2001.

\bibitem{Gross2008}
R.~Gross, I.~Matthews, and S.~Baker, ``Multi-pie,'' in \emph{IEEE Conference on
  Automatic Face and Gesture Recognition}, 2008.

\bibitem{Ahonen2006}
T.~Ahonen, A.~Hadid, and M.~Pietikainen, ``Face description with local binary
  patterns: application to face recognition,'' \emph{IEEE TPAMI}, vol.~28,
  no.~12, pp. 2037--2041, 2006.

\bibitem{Jia13caffe}
Y.~Jia, ``{Caffe}: An open source convolutional architecture for fast feature
  embedding,'' \url{http://caffe.berkeleyvision.org/}, 2013.

\bibitem{Georghiades2001}
A.~S. Georghiades, P.~N. Belhumeur, and D.~J. Kriegman, ``From few to many:
  Illumination cone models for face recognition under variable lighting and
  pose,'' \emph{IEEE TPAMI}, vol.~23, no.~6, pp. 643--660, June 2001.

\bibitem{Tan2010}
X.~Tan and B.~Triggs, ``Enhanced local texture feature sets for face
  recognition under difficult lighting conditions,'' \emph{IEEE TIP}, vol.~19,
  no.~6, pp. 1635--1650, 2010.

\bibitem{Martinez1998}
A.~Martinez and R.~Benavente, \emph{CVC Technical Report 24}, 1998.

\bibitem{Deng2012}
W.~Deng, J.~Hu, and J.~Guo, ``Extended {SRC}: Undersampled face recognition via
  intraclass variant dictionary,'' \emph{IEEE TPAMI}, vol.~34, no.~9, pp.
  1864--1870, Sept. 2012.

\bibitem{Jonathon1998}
P.~J. Phillips, H.~Wechsler, J.~Huang, and P.~J. Rauss, ``The {FERET} database
  and evaluaion procedure for face-recognition algorithms,'' \emph{Image Vision
  Comput.}, vol.~16, no.~5, pp. 295--306, 1998.

\bibitem{Lu2013}
J.~Lu, Y.-P. Tan, and G.~Wang, ``Discriminative multi-manifold analysis for
  face recognition from a single training sample per person,'' \emph{IEEE
  TPAMI}, vol.~35, no.~1, pp. 39--51, 2013.

\bibitem{Vu2012}
N.-S. Vu and A.~Caplier, ``Enhanced patterns of oriented edge magnitudes for
  face recognition and image matching,'' \emph{IEEE TIP}, vol.~21, no.~3, pp.
  1352--1368, 2012.

\bibitem{Hussain2012}
S.~Hussain, T.~Napoleon, and F.~Jurie, ``Face recognition using local quantized
  patterns,'' in \emph{BMVC}, 2012.

\bibitem{Xie2010}
S.~Xie, S.~Shan, X.~Chen, and J.~Chen, ``Fusing local patterns of gabor
  magnitude and phase for face recognition,'' \emph{IEEE TIP}, vol.~19, no.~5,
  pp. 1349--1361, 2010.

\bibitem{Vu2013}
N.-S. Vu, ``Exploring patterns of gradient orientations and magnitudes for face
  recognition,'' \emph{IEEE Trans. Information Forensics and Security}, vol.~8,
  no.~2, pp. 295--304, 2013.

\bibitem{Chai2014}
Z.~Chai, Z.~Sun, H.~Méndez-Vázquez, R.~He, and T.~Tan, ``Gabor ordinal
  measures for face recognition,'' \emph{IEEE Trans. Information Forensics and
  Security}, vol.~9, no.~1, pp. 14--26, Jan. 2014.

\bibitem{Huang2007}
G.~Huang, M.~Ramesh, T.~Berg, and E.~Learned-Miller, ``Labeled faces in the
  wild: a database for studying face recognition in unconstrained
  environments,'' in \emph{Technical Report 07-49, University of Massachusetts,
  Amherst}, 2007.

\bibitem{Wolf2011}
L.~Wolf, T.~Hassner, and Y.~Taigman, ``Effective face recognition by combining
  multiple descriptors and learned background statistics,'' \emph{IEEE TPAMI},
  vol.~33, no.~10, 2011.

\bibitem{Barkan2013}
O.~Barkan, J.~Weill, L.~Wolf, and H.~Aronowitz, ``Fast high dimensional vector
  multiplication face recognition,'' in \emph{IEEE ICCV}, 2013.

\bibitem{Taigman2014}
Y.~Taigman, M.~Yang, M.~A. Ranzato, and L.~Wolf, ``Deepface: {C}losing the gap
  to human-level performance in face verification,'' in \emph{IEEE CVPR}, 2014.

\bibitem{Fan2014}
H.~Fan, Z.~Cao, Y.~Jiang, and Q.~Yin, ``Learning deep face representation,'' in
  \emph{arXiv: 1403.2802v1}, 2014.

\bibitem{Chen2013}
D.~Chen, X.~Cao, F.~Wen, and J.~Sun, ``Blessing of dimensionality:
  high-dimentional feature and its efficient compression for face
  verification,'' in \emph{CVPR}, 2013.

\bibitem{Cui2013}
Z.~Cui, W.~Li, D.~Xu, S.~Shan, and X.~Chen, ``Fusing robust face region
  descriptors via multiple metric learning for face recognition in the wild,''
  in \emph{CVPR}, 2013.

\bibitem{Larochelle2007}
H.~Larochelle, D.~Erhan, A.~Courville, J.~Bergstra, and Y.~Bengio, ``An
  empirical evaluation of deep architectures on problems with many factors of
  variation,'' in \emph{ICML}, 2007.

\bibitem{Ciresan2012}
D.~Ciresan, U.~Meier, and J.~Schmidhuber, ``Multi-column deep neural networks
  for image classification,'' in \emph{CVPR}, 2012.

\bibitem{Sohn2013}
K.~Sohn, G.~Zhou, C.~Lee, and H.~Lee, ``Learning and selecting features jointly
  with point-wise gated {B}oltzmann machine,'' in \emph{ICML}, 2013.

\bibitem{Yu2011}
K.~Yu, Y.~Lin, and J.~Lafferty, ``Learning image representations from the pixel
  level via hierarchical sparse coding,'' in \emph{CVPR}, 2011.

\bibitem{Belongie2002}
S.~Belongie, J.~Malik, and J.~Puzicha, ``Shape matching and object recognition
  using shape contexts,'' \emph{IEEE TPAMI}, vol.~24, no.~4, pp. 509--522,
  2002.

\bibitem{Keysers2007}
D.~Keysers, T.~Deselaers, C.~Gollan, and H.~Ney, ``Deformation models for image
  recognition,'' \emph{IEEE TPAMI}, vol.~29, no.~8, pp. 1422--1435, 2007.

\bibitem{Zeiler2013}
M.~D. Zeiler and R.~Fergus, ``Stochastic pooling for regularization of deep
  convolutional neural networks,'' in \emph{ICLR}, 2013.

\bibitem{Rifai2011}
S.~Rifai, P.~Vincent, X.~Muller, X.~Glorot, and Y.~Bengio, ``Contractive
  auto-encoders: explicit invariance during feature extraction,'' in
  \emph{ICML}, 2011.

\bibitem{Sohn2012}
K.~Sohn and H.~Lee, ``Learning invariant representations with local
  transformations,'' in \emph{ICML}, 2012.

\bibitem{Varma2009}
M.~Varma and A.~Zisserman, ``A statistical approach to material classification
  using image patch examplars,'' \emph{IEEE TPAMI}, vol.~31, no.~11, pp.
  2032--2047, 2009.

\bibitem{Hayman2004}
E.~Hayman, B.~Caputo, M.~Fritz, and J.~O. Eklundh, ``On the significance of
  real-world conditions for material classification,'' in \emph{ECCV}, 2004.

\bibitem{Crosier2010}
M.~Crosier and L.~Griffin, ``Using basic image features for texture
  classification,'' \emph{IJCV}, pp. 447--460, 2010.

\bibitem{Broadhurst2006}
R.~E. Broadhust, ``Statistical estimation of histogram variation for texture
  classification,'' in \emph{Proc. Workshop on Texture Analysis and Synthesis},
  2006.

\bibitem{Grauman2005}
K.~Grauman and T.~Darrell, ``The pyramid match kernel: Discriminative
  classification with sets of image features,'' in \emph{ICCV}, 2005.

\bibitem{Lazebnik2006}
S.~Lazebnik, C.~Scmid, and J.~Ponce, ``Beyond bags of features: spatial pyramid
  matching for recognizing natural scene categories,'' in \emph{CVPR}, 2006.

\bibitem{He2014}
K.~He, X.~Zhang, S.~Ren, and J.~Sun, ``Spatial pyramid pooling in deep
  convolutional networks for visual recognition,'' in \emph{ECCV}, 2014.

\bibitem{Le2010}
Q.~V. Le, J.~Ngiam, Z.~Chen, D.~Chia, P.~W. Koh, and A.~Y. Ng, ``Tiled
  convolutional neural networks,'' in \emph{NIPS}, 2010.

\bibitem{Yu2010}
K.~Yu and T.~Zhang, ``Improved local coordinate coding using local tangents,''
  in \emph{ICML}, 2010.

\bibitem{Bo2010}
L.~Bo, X.~Ren, and D.~Fox, ``Kernel descriptors for visual recognition,'' in
  \emph{NIPS}, 2010.

\bibitem{Coates2010}
A.~Coates, H.~Lee, and A.~Y. Ng, ``An analysis of single-layer networks in
  unsupervised feature learning,'' in \emph{NIPS Workshop}, 2010.

\bibitem{Krizhevsky2014}
A.~Krizhevsky, ``cuda-convnet,'' \url{http://code.google.com/p/cuda-convnet/},
  {J}uly 18, 2014.

\bibitem{Snoek2012}
J.~Snoek, H.~Larochelle, and R.~P. Adams, ``Practical bayesian optimization of
  machine learning algorithms,'' in \emph{NIPS}, 2012.

\bibitem{Lin2014}
M.~Lin, Q.~Chen, and S.~Yan, ``Network in network,'' in
  \emph{arXiv:1312.4400v3}, 2014.

\end{thebibliography}
%}

%
\begin{IEEEbiography}[{\includegraphics[width=1in,height=1.25in,clip,keepaspectratio]{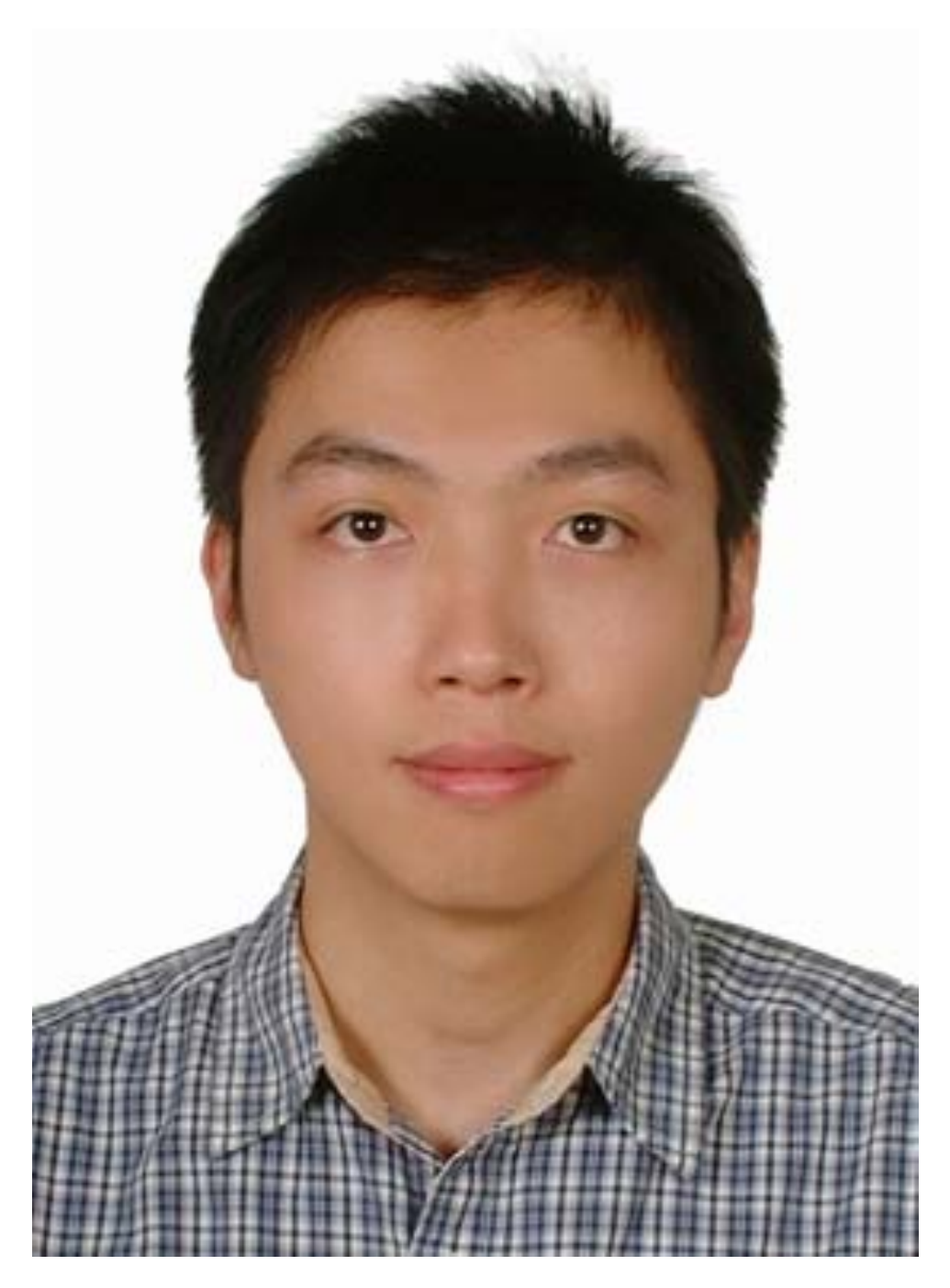}}]
{Tsung-Han Chan} received the B.S. degree from the Department of Electrical Engineering, Yuan Ze University, Taiwan, in 2004
and the Ph.D. degree from the Institute of Communications Engineering, National Tsing Hua University, Taiwan, in 2009. He is currently working as a Project Lead R$\&$D Engineer with Sunplus Technology Co., Hsinchu, Taiwan. His research interests are in image processing and convex optimization, with a recent emphasis on computer vision and hyperspectral remote sensing.
\end{IEEEbiography}

\begin{IEEEbiography}[{\includegraphics[width=1in,height=1.25in,clip,keepaspectratio]{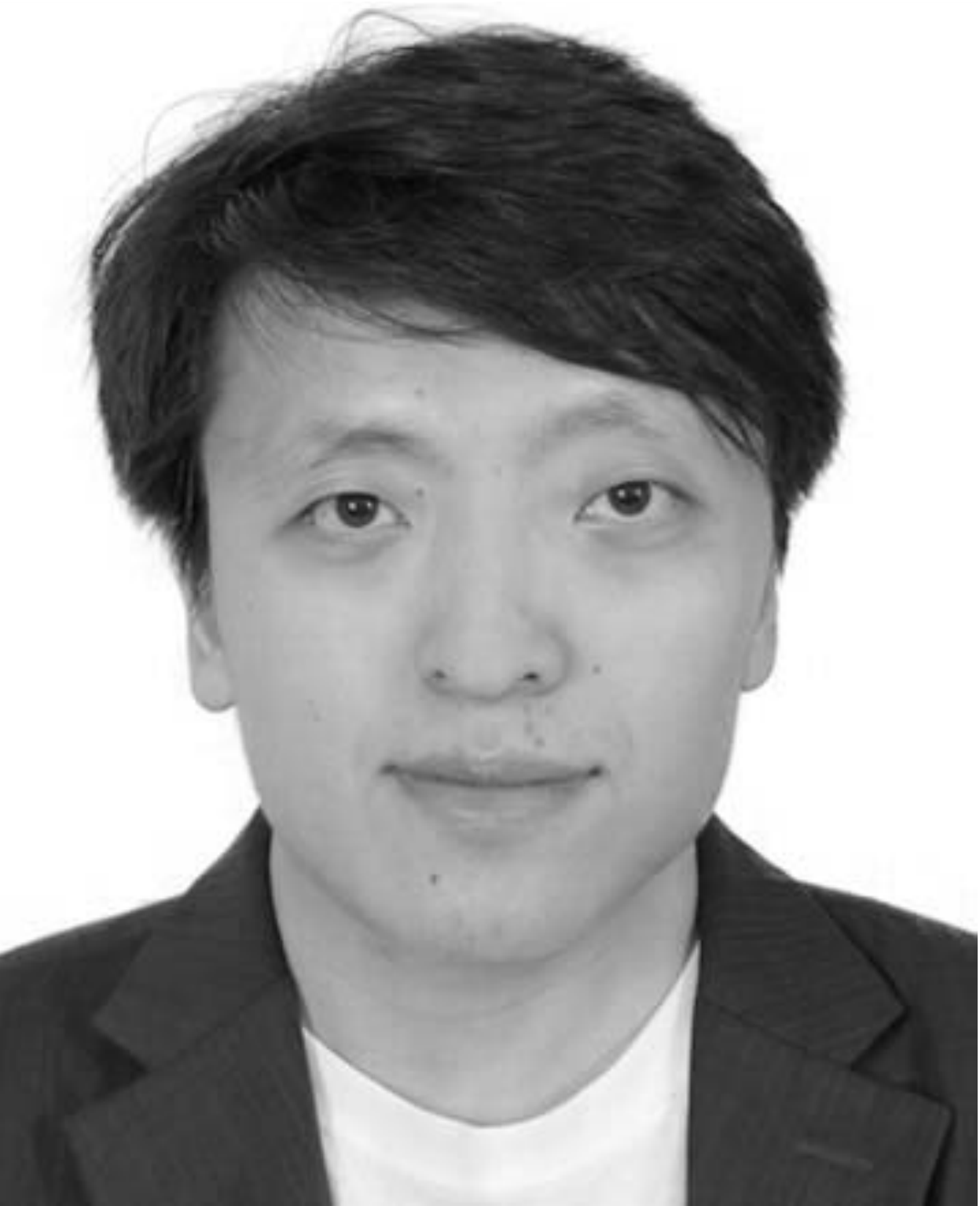}}]
{Kui Jia} received the B.Eng. degree in marine engineering from Northwestern Polytechnical University, China, in 2001, the M.Eng. degree in electrical and computer engineering from National University of Singapore in 2003, and the Ph.D. degree in computer science from Queen Mary, University of London, London, U.K., in 2007. He is currently a Research Scientist at Advanced Digital Sciences Center. His research interests are in computer vision, machine learning, and image processing.
\end{IEEEbiography}

\begin{IEEEbiography}[{\includegraphics[width=1in,height=1.25in,clip,keepaspectratio]{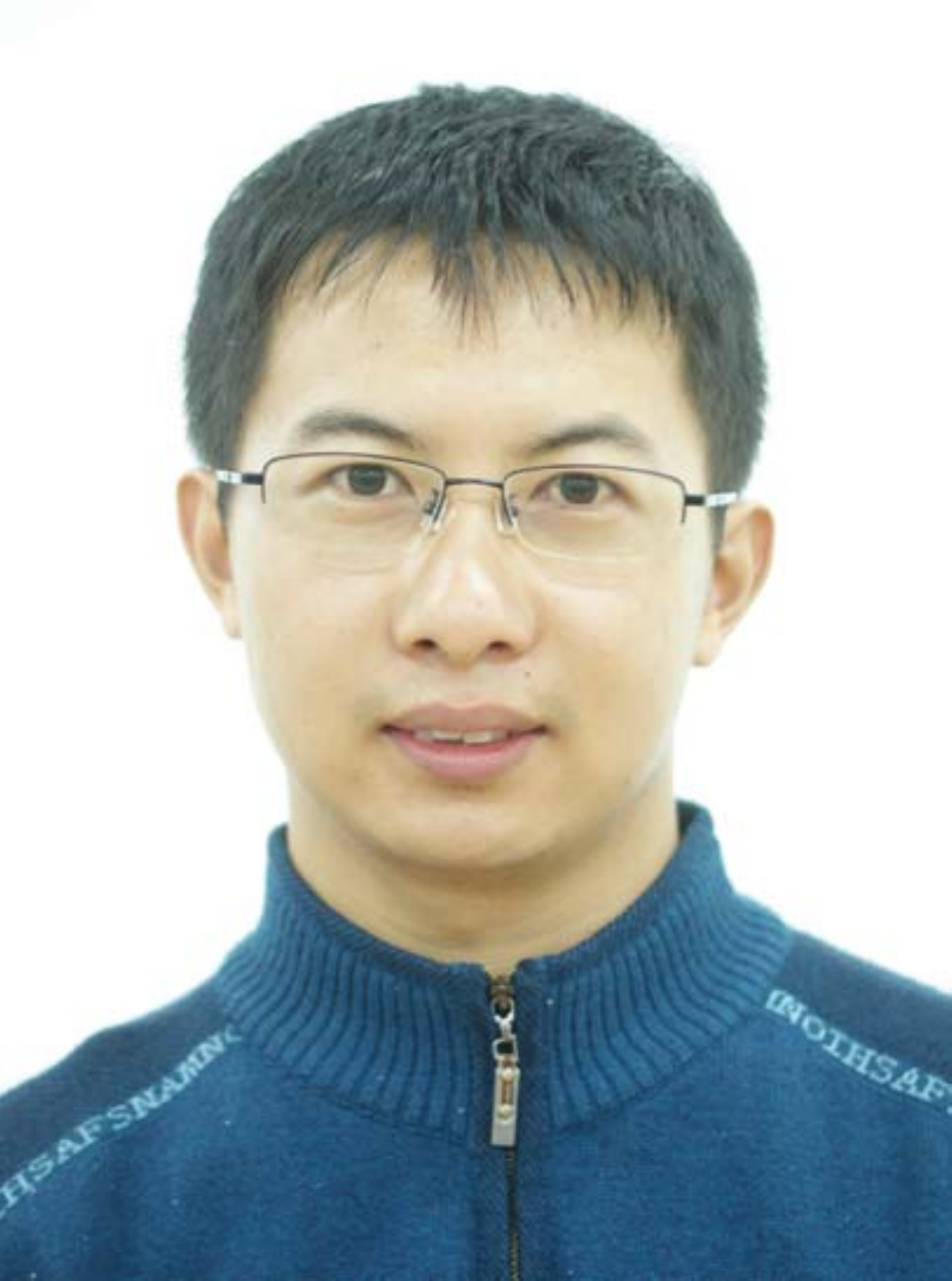}}]
{Shenghua Gao} received the B.E. degree from the University of Science and Technology of China in 2008, and received the Ph.D. degree from the Nanyang Technological University in 2013. He is currently a postdoctoral fellow in Advanced Digital Sciences Center, Singapore. He was awarded the Microsoft Research Fellowship in 2010. His research interests include computer vision and machine learning.
\end{IEEEbiography}

\begin{IEEEbiography}[{\includegraphics[width=1in,height=1.25in,clip,keepaspectratio]{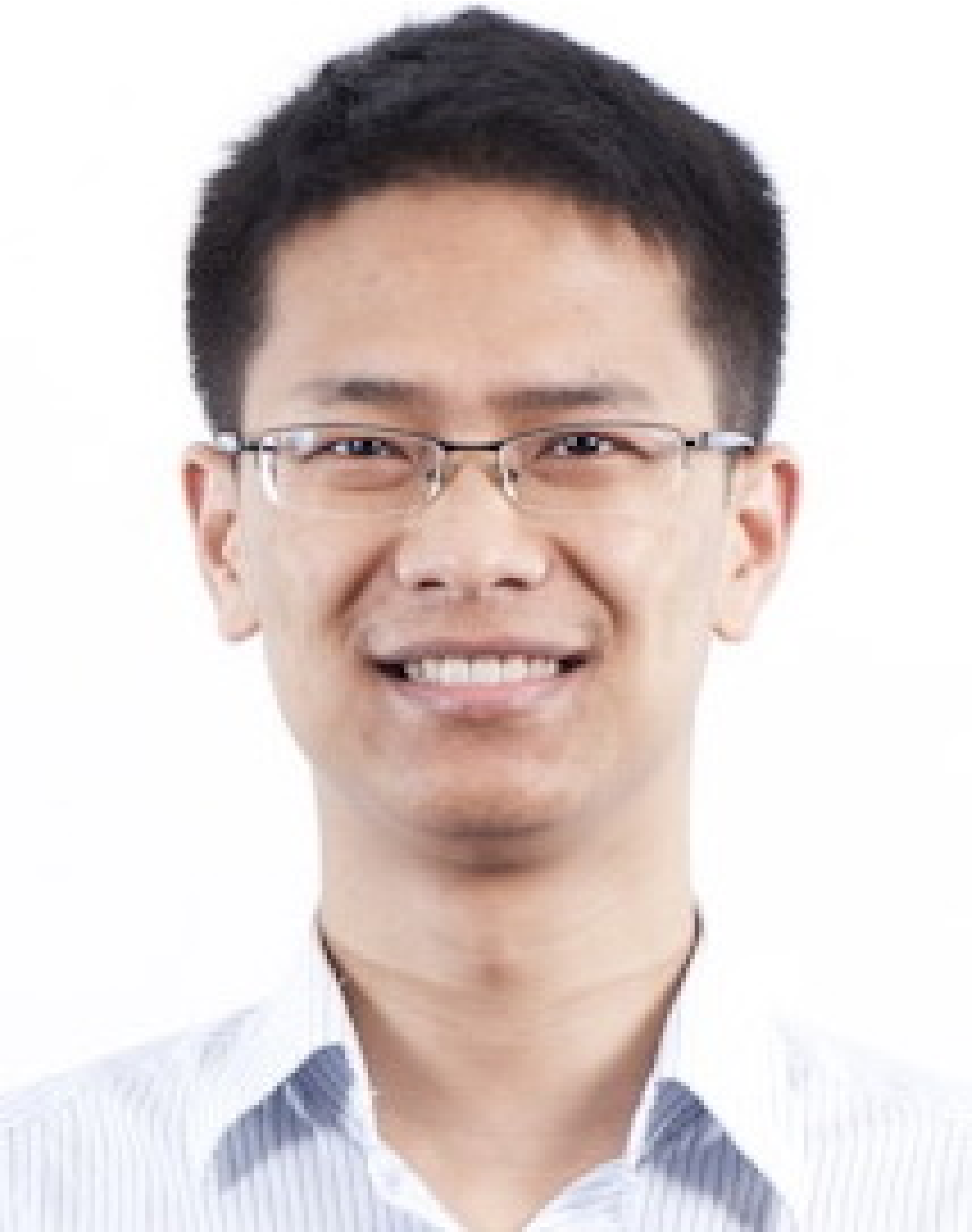}}]
{Jiwen Lu} is currently a research scientist at the Advanced Digital Sciences Center (ADSC), Singapore. His research interests include computer vision, pattern recognition, machine learning, and biometrics. He has authored/co-authored more than 70 scientific papers in peer-reviewed journals and conferences including some top venues such as the TPAMI, TIP, CVPR and ICCV. He is a member of the IEEE.
\end{IEEEbiography}

\begin{IEEEbiography}[{\includegraphics[width=1in,height=1.25in,clip,keepaspectratio]{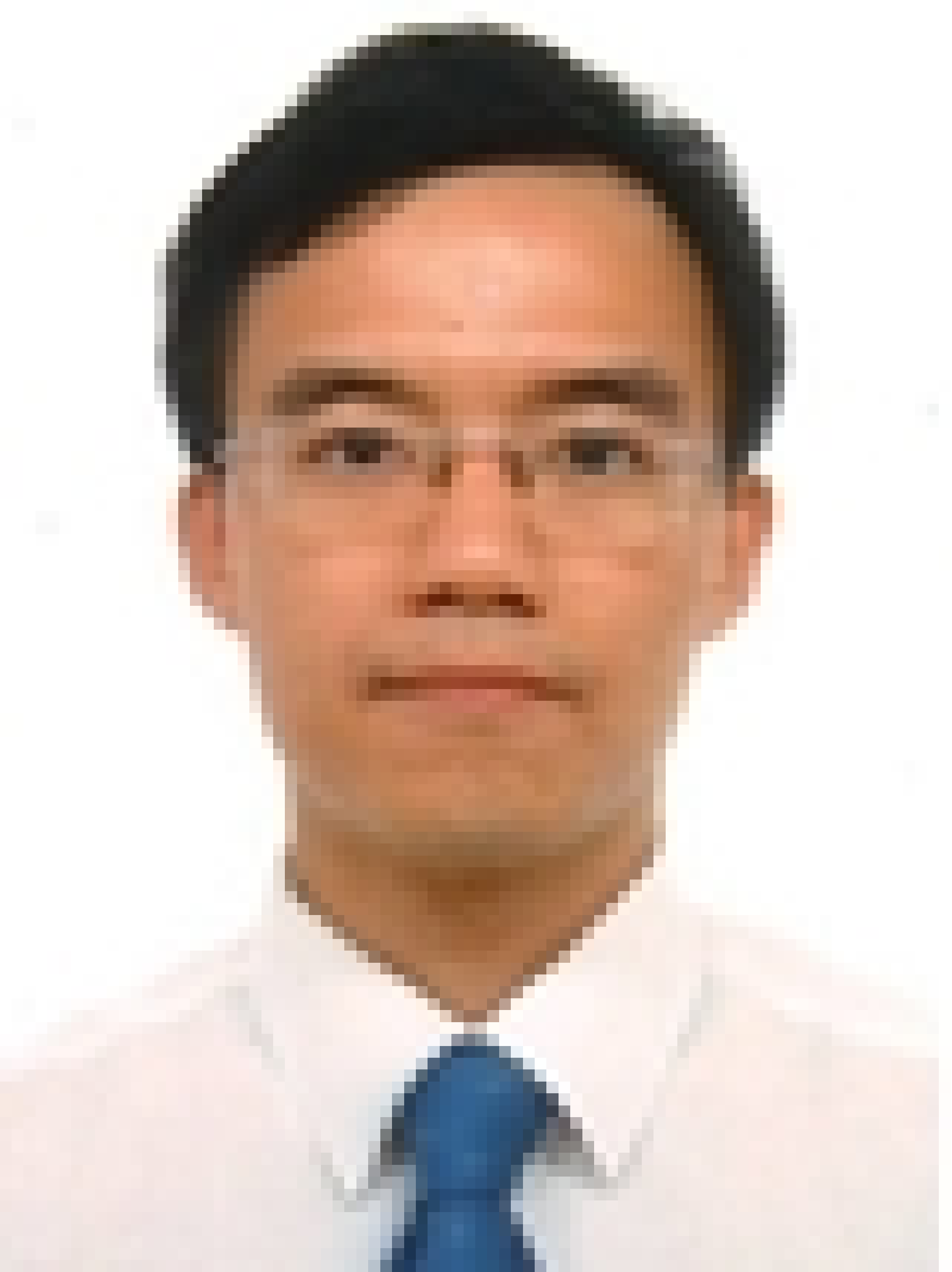}}]
{Zinan Zeng} received the master degree and B.E. degree with First Honour from the School of Computer Engineering, Nanyang Technological University, Singapore. He is now a senior software engineer in Advanced Digital Sciences Center, Singapore. His research interests include statistical learning, optimization with application in computer vision.
\end{IEEEbiography}

\begin{IEEEbiography}[{\includegraphics[width=1in,height=1.25in,clip,keepaspectratio]{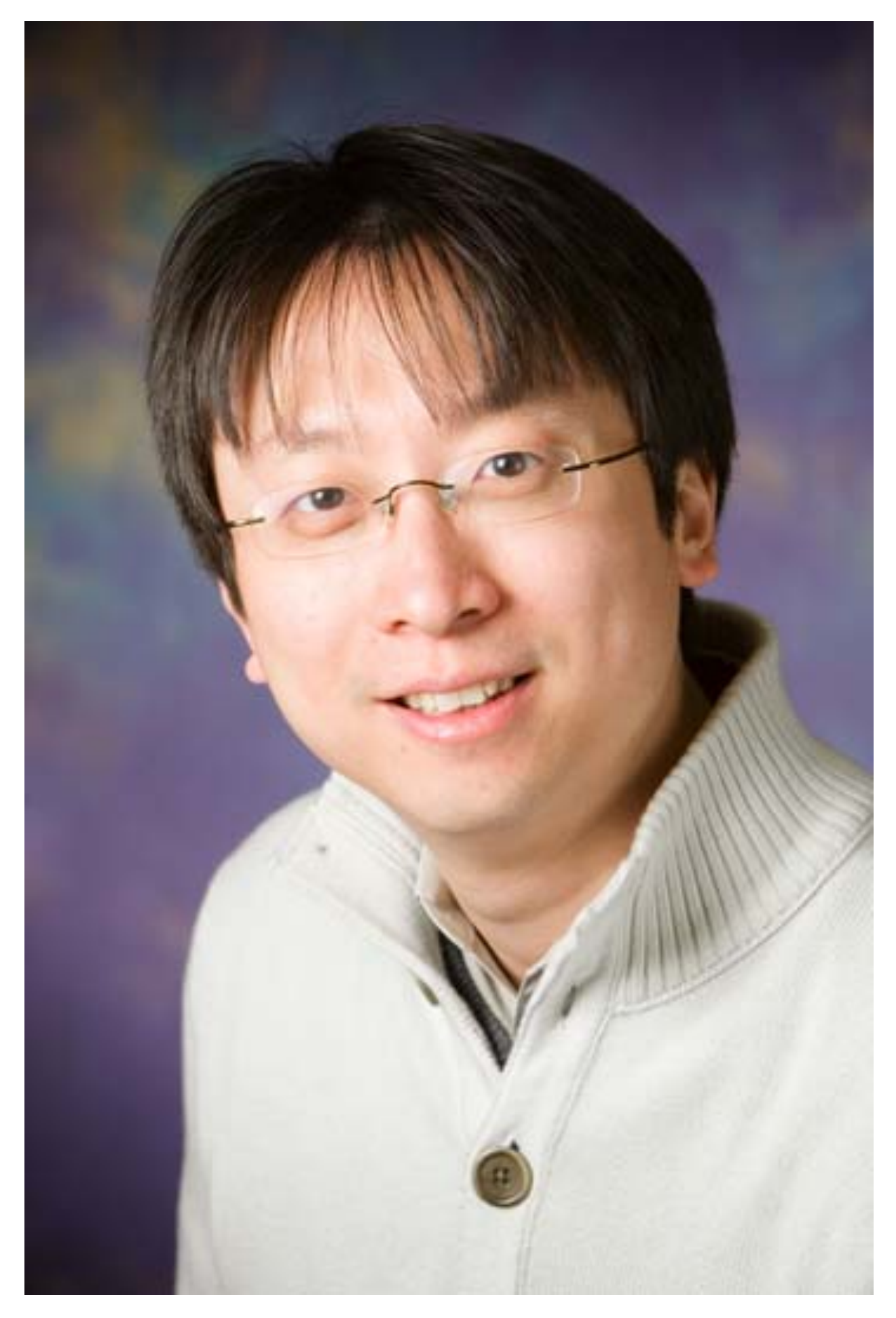}}]
{Yi Ma}(F'13) is a  professor of the School of Information Science and
Technology of ShanghaiTech University. He received his Bachelors' degree in Automation and Applied
Mathematics from Tsinghua University, China in 1995. He
received his M.S. degree in EECS in 1997, M.A. degree in
Mathematics in 2000, and his PhD degree in EECS in 2000 all from
UC Berkeley. From 2000 to 2011, he was an associate professor of the ECE
Department of the University of Illinois at Urbana-Champaign, where he now holds an adjunct position.
From 2009 to early 2014, he was a principal researcher and manager of the visual computing
group of Microsoft Research Asia. His main research areas are in computer vision and high-dimensional data analysis.
Yi Ma was the recipient of the David Marr Best Paper Prize from ICCV 1999 and Honorable Mention for the Longuet-Higgins Best
Paper Award from ECCV 2004. He received the CAREER Award from the National Science Foundation
in 2004 and the Young Investigator Program Award from the
Office of Naval Research in 2005. He has been an associate editor for IJCV, SIIMS, IEEE Trans. PAMI and Information Theory.

\end{IEEEbiography}

\end{document}